\documentclass[lettersize,compsoc,journal]{IEEEtran}
\usepackage{amsmath,amsfonts}
\usepackage{algorithm}
\usepackage{array}
\usepackage[caption=false,font=normalsize,labelfont=sf,textfont=sf]{subfig}
\usepackage{textcomp}
\usepackage{stfloats}
\usepackage{url}
\usepackage{verbatim}
\usepackage{graphicx}
\usepackage{cite}
\usepackage{times}
\usepackage{latexsym}
\usepackage{booktabs}
\usepackage{siunitx}

\usepackage{cite}
\usepackage{algorithm}
\usepackage{algpseudocode}%
\usepackage{amsmath}
\usepackage{amsmath} 
\usepackage{amssymb}
\usepackage{url}
\usepackage{breakurl}
\usepackage{hyperref}

\usepackage[misc]{ifsym}
\usepackage{bm}
\usepackage{booktabs}
\usepackage{makecell}  
\usepackage{array,color}
\usepackage{array}
\usepackage{multirow}

\usepackage[misc]{ifsym}
\usepackage{multirow}  
\usepackage{booktabs}
\usepackage{threeparttable}
\usepackage{bbding}
\usepackage{ulem}
\newcommand{\etal}{\emph{et al.}}

\usepackage{pifont}
\usepackage{tabularx}

\begin{document}

\title{TopoSculpt: Betti-Steered Topological Sculpting of Three-Dimensional Tubular Shapes}

\author{Minghui Zhang, Yaoyu Liu, Junyang Wu, Xin You, Hanxiao Zhang, Junjun He, Yun Gu,~\IEEEmembership{Member,~IEEE}, Xinglin Zhang
\thanks{Minghui Zhang (e-mail:minghuizhang@sjtu.edu.cn), Yaoyu Liu, Junyang Wu, Xin You, Hanxiao Zhang and Yun Gu are with Institute of Medical Robotics, Shanghai Jiao Tong University, Shanghai, China. \\
Minghui Zhang, Yaoyu Liu, Xin You and Yun Gu are also with Department of Automation, Shanghai Jiao Tong University, Shanghai, China. \\
Junjun He is with Shanghai AI Lab, Shanghai, China. \\
Xinglin Zhang is with Medical Image Insights, Shanghai, China. \\}
}

\markboth{Journal of \LaTeX\ Class Files,~Vol.~14, No.~8, August~2021}%
{Zhang \MakeLowercase{\textit{et al.}}: A Sample Article Using IEEEtran.cls for IEEE Journals}


\IEEEtitleabstractindextext{
\begin{abstract}
Medical tubular anatomical structures are inherently three-dimensional conduits with lumens, enclosing walls, and complex branching topologies. Accurate reconstruction of their geometry and topology is crucial for applications such as bronchoscopic navigation and cerebral arterial connectivity assessment. Existing methods often rely on voxel-wise overlap measures, which fail to capture topological correctness and completeness. Although topology-aware losses and persistent homology constraints have shown promise, they are usually applied patch-wise and cannot guarantee global preservation or correct geometric errors at inference.
To address these limitations, we propose a novel TopoSculpt, a framework for topological refinement of 3D fine-grained tubular structures. TopoSculpt (i) adopts a holistic whole-region modeling strategy to capture full spatial context, (ii) first introduces a Topological Integrity Betti (TIB) constraint that jointly enforces Betti number priors and global integrity, and (iii) employs a curriculum refinement scheme with persistent homology to progressively correct errors from coarse to fine scales.
Extensive experiments on three challenging tubular datasets, including the pulmonary airway, the Circle of Willis (CoW), and the coronary artery (CTCA), demonstrate substantial improvements in both geometric accuracy and topological fidelity. For instance, the $\beta_{0}$ error is reduced from 69.00 to 3.40 on the airway dataset, from 1.65 to 0.30 on the CoW dataset, and from 8.75 to 0.88 on the CTCA dataset, with Tree Detection (TD) and Branch Detection (BD) rates improving by approximately 10\%. These results highlight the effectiveness of TopoSculpt in correcting critical topological errors and advancing high-fidelity modeling of complex 3D tubular anatomy.
The project homepage is available at: \url{https://github.com/EndoluminalSurgicalVision-IMR/TopoSculpt}.
\end{abstract}

}
\maketitle

\section{Introduction}

\IEEEPARstart{I}{n} medical anatomy, tubular structures represent a fundamental class of organs characterized by a lumen enclosed by specialized walls, serving as conduits for the transport of gases, fluids, or other biological materials \cite{sutton1987textbook, brunicardi2014schwartz}. For instance, the pulmonary airway~\cite{zhang2023multi} is a key component of the respiratory system and functions primarily in the conduction of air, whereas the Circle-of-Willis (CoW)\cite{yang2024benchmarking} is an important anastomotic network of arteries connecting the anterior and posterior circulations of the brain.
Importantly, tubular structures are inherently three-dimensional in their physical organization. Precise characterization of their spatial geometry and topology is essential for elucidating physiological functions, detecting pathological alterations, and guiding clinical decision-making and interventions.

\begin{figure}[t]
\centering
\includegraphics[width=1.00\linewidth]{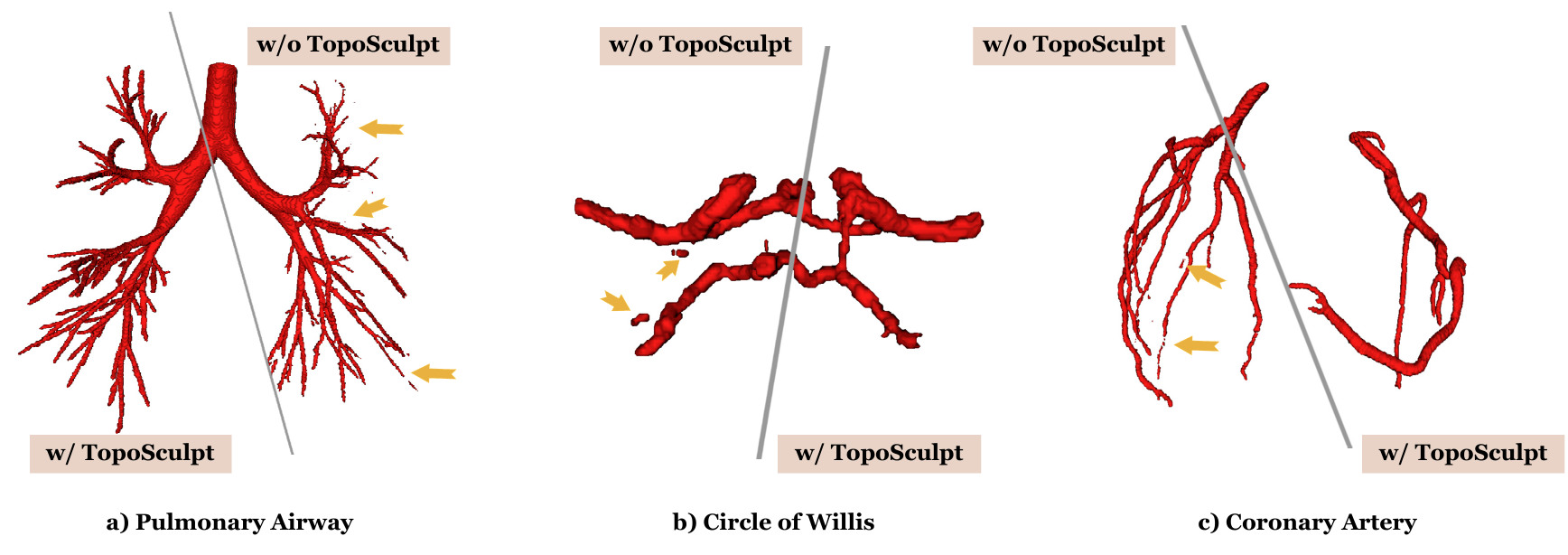}
\caption{
Existing methods often yield fragmented or disconnected structures (arrows) when modeling fine-grained three-dimensional tubular anatomy, whereas the proposed TopoSculpt improves structural continuity and preserves global topological integrity (from left to right: pulmonary airway, Circle of Willis, and coronary artery).
}
\label{fig:TopoSculpt_Glance}
\end{figure}

Medical tubular structures differ from line-based structures such as those in road extraction datasets (e.g., DeepGlobe \cite{demir2018deepglobe}, Massachusetts Roads \cite{mnih2013machine}, and CNDS \cite{cheng2017automatic}). 
Unlike existing road datasets, which are typically modeled as two-dimensional line features without thickness, tubular anatomical structures are volumetric conduits with lumens, enclosing walls, and complex branching topologies. Their clinical significance critically depends on accurate reconstruction of both geometry and connectivity.
Furthermore, anatomical tubular structures exhibit consistent topological similarity across individuals: even when variations occur, these differences are clinically meaningful and can be systematically studied as anatomical variability. In contrast, the topology of 2D road networks is largely image-dependent and spatially arbitrary, lacking the inherent cross-sample consistency that defines biological tubular systems. 
Moreover, many widely-used retinal vessel imaging modalities remain two-dimensional in nature, such as fundus photography datasets (e.g., DRIVE \cite{staal2004ridge}, STARE \cite{hoover2000locating}, CHASEDB1 \cite{fraz2012ensemble}) and OCT-angiography datasets (e.g., ROSE \cite{ma2020rose}, OCTA-500 \cite{li2024octa}). While these datasets have greatly facilitated progress in retinal vessel analysis, they inherently provide only a planar projection of the vasculature and thus may not fully capture the authentic three-dimensional architecture and topological intricacies of fine-scale vascular networks. In this paper, tubular structures are intrinsically three-dimensional, making the accurate reconstruction of their 3D geometry and topology indispensable for identifying pathological alterations and guiding precise clinical interventions. Preserving the topological integrity of these structures during segmentation is particularly crucial, as even minor topological errors (e.g., breakage) can lead to significant misinterpretations of anatomical connectivity and function, as illustrated in Fig.\ref{fig:TopoSculpt_Glance}. For instance, in airway segmentation, only the largest connected component is clinically useful for bronchoscopic-assisted surgery \cite{li2025reflecting, gu2022vision}. Similarly, breakage compromises the reliability of connectivity analysis in cerebral arteries \cite{yang2024benchmarking}.
Moreover, preserving the topology of volumetric tubular structures is more challenging than that of 2D structures, as the complexity and connectivity of 3D shapes are inherently more intricate\cite{zhang2023multi,yang2024benchmarking}.

\begin{figure*}[t]
\centering
\includegraphics[width=1.00\linewidth]{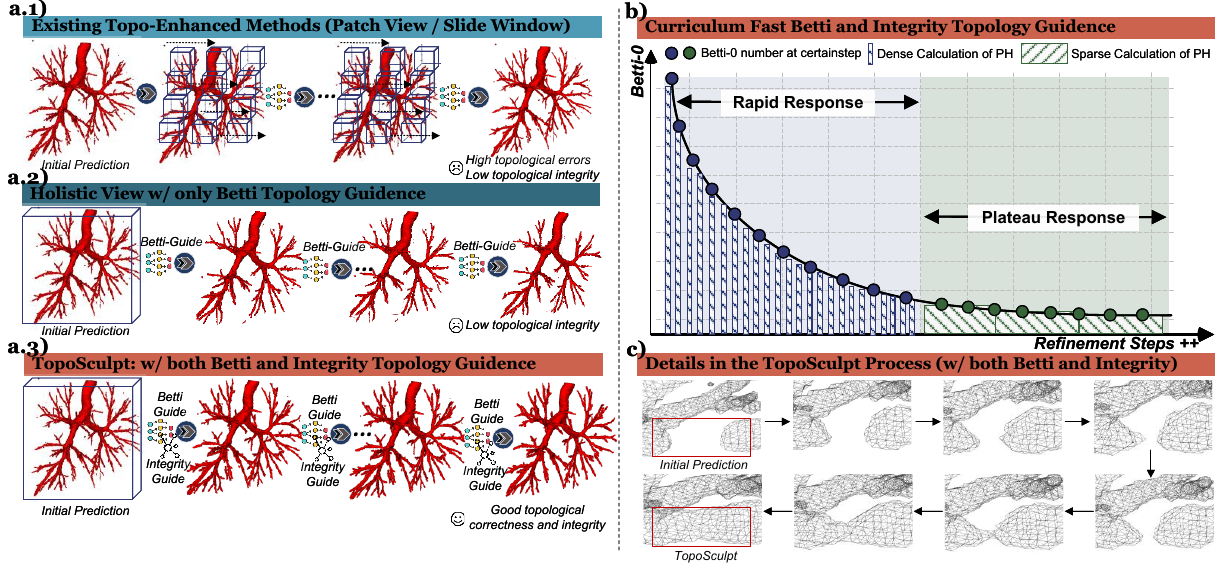}
\caption{
Pipeline of the proposed TopoSculpt.
a.1) Existing topological refinement methods typically rely on patch-wise training and sliding-window testing, which cannot guarantee the topological integrity of the whole structure, as a single patch fails to capture global geometric attributes.
a.2) In contrast, our method adopts a holistic modeling strategy by feeding the entire image region covering the complete target structure into the network, ensuring that the full spatial context is preserved. Betti-guided refinement alone can improve certain topological characteristics but often compromises global integrity.
a.3) By jointly incorporating Betti guidance and topological integrity constraints, our method preserves global topology while enhancing local topological correctness.
b) We further design a curriculum learning strategy to progressively refine segmentation results: initially correcting large-scale errors through dense persistent homology (PH) calculations, and gradually focusing on finer-scale errors with sparser PH sampling. This enables efficient learning of topological corrections across multiple scales.
c) The refinement process is supervised by both Betti numbers and topological integrity metrics, ensuring optimization consistently improves both local correctness and global fidelity.
}
\label{fig:TopoSculpt_Framework}
\end{figure*}

Unlike conventional voxel-wise segmentation tasks for large organs, where the Dice Similarity Coefficient (DSC) \cite{milletari2016v} serves as the dominant evaluation metric and its differentiable form can be directly applied as a loss function, fine-grained tubular structures present additional challenges, such as the topological correctness and completeness \cite{zhang2023multi}. Consequently, reliance solely on the DSC is insufficient for accurately modeling and evaluating the geometric fidelity and topological integrity of such complex structures. 
\textit{Accurately modeling and evaluating 3D fine-grained tubular structures remains a critical research challenge}. Such shapes can be described as envelopes of spheres or disks with continuously varying centerline points and radii \cite{benmansour2011tubular, antiga2003computational}, while their topology can be quantified by Betti numbers \cite{otter2017roadmap,wasserman2018topological}. In this work, we propose comprehensive criteria for selecting evaluation metrics that account for both overlap-based geometric accuracy and topology-based structural integrity. 


Extensive efforts have been dedicated to enhancing the topological fidelity of 3D fine-grained tubular structure segmentation through the design of topology-aware surrogate loss functions \cite{qin2019airwaynet, zheng2021alleviating, zhang2023towards, shit2021cldice, kirchhoff2024skeleton, shi2024centerline}. Such loss functions encourage the preservation of critical geometric features, including local connectivity and centerline continuity. In parallel, approaches based on Persistent Homology (PH) \cite{otter2017roadmap, zomorodian2005topology} have been explored to directly promote topological correctness of segmentation results \cite{hu2019topology,clough2020topological,hu2021topology,berger2024topologically,stucki2023topologically,hu2022learning, li2025topology}. PH-based methods operate by identifying critical points in the probability map that correspond to topological events (e.g., the creation or removal of components) and adjusting their values to better align the predicted topology with that of the ground truth (GT).
Despite these advances, several challenges remain. First, due to the high dimensionality and complexity of 3D fine-grained tubular anatomy, both surrogate loss and PH-based strategies are typically applied in a patch-wise manner. Such localized optimization may not guarantee the global preservation of topology, particularly for anatomical features extending across multiple patches. Second, topological constraints are generally imposed only during training. Once training concludes, the model parameters are fixed, leaving the network unable to correct topological inconsistency on a case-by-case basis during inference, as illustrated in Fig. ~\ref{fig:TopoSculpt_Framework}.a.1). This limitation suggests that such methods may not fully exploit the assumption that topological priors inherent in test datasets can serve as valuable guidance during the inference stage. 

To deal with these challenges, we propose a novel framework, TopoSculpt, for the topological refinement of 3D fine-grained tubular structure modeling, as demonstrated in Fig.\ref{fig:TopoSculpt_Glance}. The contributions of TopoSculpt are threefold:
First, we introduce a holistic-view modeling strategy. It is observed that complex volumetric structures in the voxel domain, such as airways, can be concisely characterized from a geometric perspective (e.g., the airway generally satisfies the attribute $\beta_{0}=1$). Building on this observation, we replace the conventional patch-wise training and sliding-window testing paradigm with a whole-region approach. Specifically, we feed entire image regions covering the complete target structures into the neural network $\mathcal{F}$, ensuring that the model captures the full spatial context of the input. This holistic perspective is critical for accurately determining the true topological characteristics of 3D shapes, both with respect to the ground truth during training and the predictions during inference. 

Second, TopoSculpt introduces a novel topological constraint that combines Betti number guidance with topological integrity preservation, referred to as Topological Integrity Betti (TIB). TIB is enforced on the basis of the holistic-view modeling strategy, since only by considering the entire structure can its global topology be accurately computed and preserved.
After $\mathcal{F}$ is trained, it generates initial predictions on unseen test images, which are then iteratively refined. The Betti number guidance encourages the segmentation to conform to known topological priors (e.g., $\beta_{0}=1$ for airways). However, enforcing Betti number constraints alone may induce undesired structural changes and undermine the topological integrity, as shown in Fig.~\ref{fig:TopoSculpt_Framework}.a.2). To mitigate this, we introduce a topological integrity term that penalizes deviations from the original topology of the predicted segmentation. By jointly optimizing these two terms, TopoSculpt preserves the global topology of the entire structure while enhancing local topological correctness, as illustrated in Fig.~\ref{fig:TopoSculpt_Framework}.a.3) and detailed in Fig.~\ref{fig:TopoSculpt_Framework}.c).

Third, we propose a curriculum refinement strategy to efficiently implement TopoSculpt. It is well known that persistent homology (PH) calculation for whole 3D structures is computationally expensive \cite{otter2017roadmap, zomorodian2005topology}. To mitigate this, we design a progressive refinement paradigm that gradually shifts the focus from large-scale to fine-scale topological errors, thereby substantially reducing the computational burden. As illustrated in Fig.~\ref{fig:TopoSculpt_Framework}.b), Betti numbers change rapidly in the early refinement stages but stabilize in later stages. Accordingly, we initially compute PH features densely with fewer steps to correct large-scale errors, and then progressively increase the sampling interval to concentrate on finer-scale corrections. This curriculum design not only accelerates the refinement process but also enables the model to efficiently learn to correct topological inaccuracies at multiple scales.

TopoSculpt was evaluated on three challenging 3D fine-grained tubular structures: the pulmonary airway \cite{zhang2023multi}, the Circle of Willis (CoW) \cite{yang2024benchmarking}, and the coronary arteries from the CTCA dataset \cite{gharleghi2023annotated}. Experimental results demonstrate that TopoSculpt significantly enhances both topological integrity and geometric accuracy across all datasets. For the pulmonary airway dataset, TopoSculpt reduces the $\beta_{0}$ error from 69.00 to 3.40, while improving the Tree Detection Rate (TD) from 82.23\% to 91.56\% and the Branch Detection Rate (BD) from 82.06\% to 91.33\%. Additionally, clDice and NSDice increase from 90.48\% to 92.27\% and from 96.96\% to 98.11\%, respectively. Similar improvements are observed on the CoW and CTCA datasets.
For the CoW dataset, which has a relatively lower initial $\beta_{0}$ error, TopoSculpt further refines the topology, reducing the $\beta_{0}$ error from 1.65 to 0.30. This correction of critical connectivity errors translates into substantial gains in TD and BD (from 82.84\% and 84.26\% to 96.73\% and 96.72\%, respectively). Concurrently, clDice and NSDice increase from 96.67\% to 97.98\% and from 99.10\% to 99.26\%, respectively.
For the CTCA dataset, TopoSculpt demonstrates robustness by reducing the $\beta_{0}$ error from 8.75 to 0.88, resulting in notable improvements in both TD (from 78.98\% to 91.68\%) and BD (from 75.17\% to 91.33\%). These consistent advancements across airway, CoW, and coronary anatomies highlight the efficacy of TopoSculpt in preserving global topological integrity while ensuring precise geometric fidelity in fine-grained 3D tubular reconstruction.

\section{Related Works}\label{sec:related_works}
\subsection{Evaluation Metrics for Tubular Shapes}
For general medical anatomy segmentation, the Dice Similarity Coefficient (DSC) ~\cite{milletari2016v} is the most commonly used evaluation metric, and its differentiable form is often adopted as a loss function. In contrast, three-dimensional tubular structures with complex branching patterns, varying radii, and topology-sensitive connectivity pose additional challenges, where DSC alone cannot adequately capture geometric fidelity or topological integrity ~\cite{zhang2023multi}. 
Recent studies ~\cite{reinke2021common,maier2024metrics} have emphasized the limitations of overlap-based measures and suggested geometry-aware alternatives. The centerline Dice (clDice) ~\cite{shit2021cldice} and normalized surface Dice (NSDice) ~\cite{nikolov2018deep} jointly evaluate centerline continuity and boundary accuracy, while the Hausdorff Distance (HD) ~\cite{huttenlocher2002comparing} captures the maximum boundary deviation. Furthermore, Betti numbers~\cite{otter2017roadmap,wasserman2018topological} are employed to quantify topological correctness by evaluating the number of connected components, while the Tree Length Detected (TD) and Branch Detected Rate (BD)~\cite{zhang2023multi} measure topological completeness by assessing the recovery of major branches.   
We emphasize that these topology-oriented metrics are critical for assessing the modeling quality of medical tubular structures, as they better represent the true anatomical structure and overall connectivity than conventional overlap-based metrics.

\subsection{Topology-aware Segmentation}
Recent studies have proposed various topology-aware loss functions to better preserve the topologically critical regions of tubular structures.  
Kervadec \etal ~\cite{kervadec2019boundary} introduced a boundary loss formulated as a distance metric in the contour space to improve segmentation accuracy along object boundaries.  
The centerline Dice (clDice) loss ~\cite{shit2021cldice} was designed to explicitly enforce connectivity preservation in thin structure segmentation, and Shi \etal ~\cite{shi2024centerline} extended it to a centerline-boundary Dice (cbDice) loss by incorporating boundary-aware components.  
However, the differentiable skeletonization process in clDice tends to produce jagged or unstable centerlines. To address this issue, Kirchhoff \etal~\cite{kirchhoff2024skeleton} and Zhang \etal ~\cite{zhang2023towards} employed more robust skeletonization strategies and developed skeleton-based weighted loss functions to further enhance topological accuracy. 
In parallel, Persistent Homology (PH) ~\cite{otter2017roadmap} based approaches have been explored to directly enforce topological correctness in segmentation results ~\cite{hu2019topology,hu2021topology}.  
Hu \etal ~\cite{hu2019topology} introduced a topological loss that penalizes discrepancies between the persistence diagrams of predictions and ground truth. Subsequently, Discrete Morse theory was incorporated into PH ~\cite{hu2021topology} to localize critical points and further improve the topological accuracy of segmentation. 
Despite these advances, several challenges remain. Owing to the high dimensionality and structural complexity of 3D fine-grained tubular anatomy, both surrogate loss and PH-based methods are often applied in a patch-wise manner, which limits their ability to ensure global topological consistency. In addition, PH computation is computationally expensive for high-dimensional data, reducing its practicality for large 3D volumes.

\section{Method}

\begin{figure*}[t]
\centering
\includegraphics[width=1.00\linewidth]{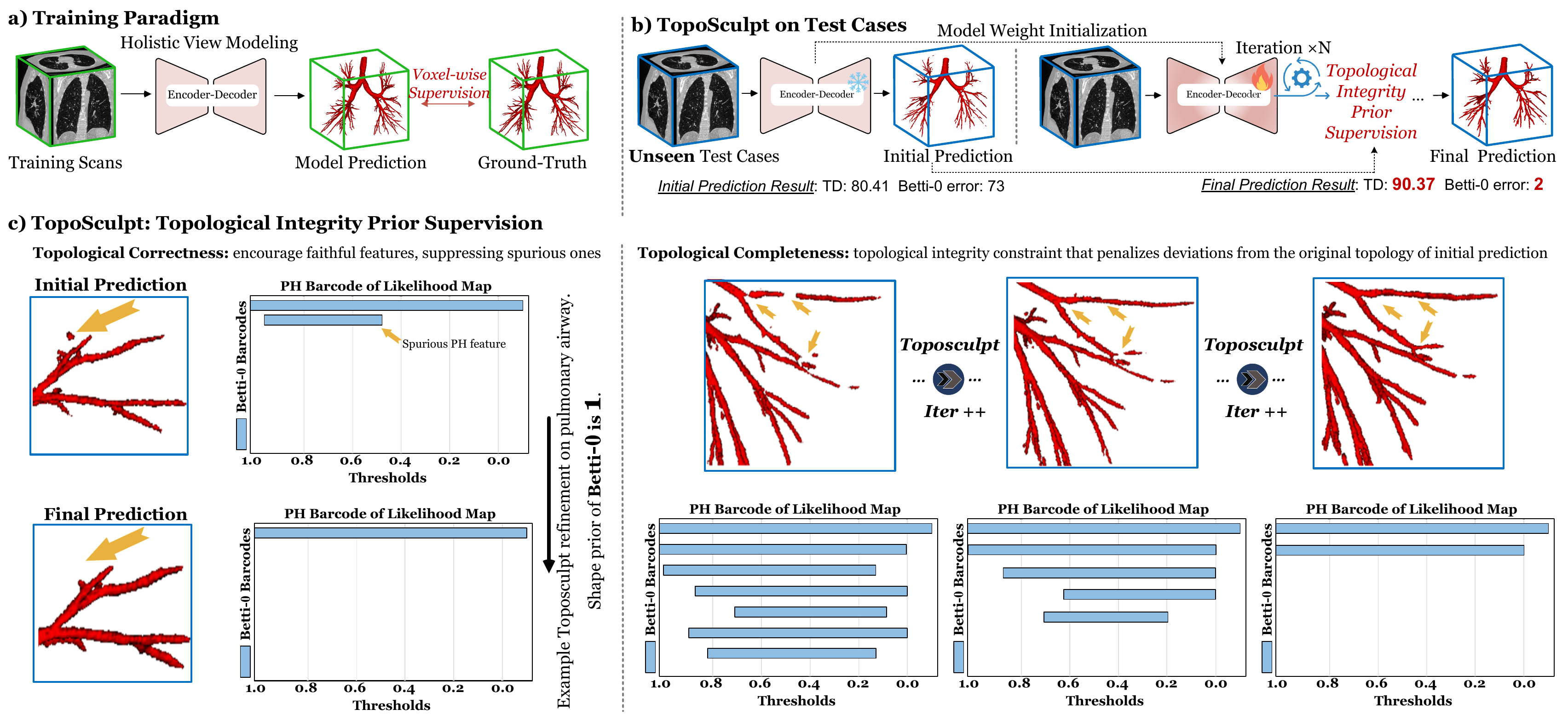}
\caption{
Detailed framework of the proposed TopoSculpt. 
(a) Training paradigm. TopoSculpt is optimized via a holistic-view modeling strategy, where entire anatomical regions encompassing the complete target structure are fed into the network to capture full spatial and topological context. Voxel-wise supervision is applied in the training stage.
(b) TopoSculpt refinement on unseen test cases. For a given test scan, the initial prediction is generated by the trained encode-decoder network and iteratively refined under the supervision of a topological integrity prior. This refinement progressively corrects connectivity errors, thereby improving the topological accuracy.
(c) Details of the topological integrity prior supervision. During refinement, the persistent homology (PH) barcodes of the likelihood maps are analyzed to identify critical topological features. The Betti-guided correction term encourages the removal of spurious PH components (enhancing topological correctness), while the topological integrity constraint penalizes deviations from the original global topology, preserving topological completeness. Through iterative refinement, TopoSculpt achieves consistent suppression of erroneous components and convergence toward topologically faithful 3D reconstructions. 
}
\label{fig:TopoSculpt_Pipeline}
\end{figure*}

\subsection{Holistic View Modeling}
As illustrated in Fig.\ref{fig:TopoSculpt_Pipeline}.a), instead of conventional patch-wise training and sliding-window testing, we employ a whole-region strategy by feeding entire image regions containing complete target structures into the network $\mathcal{F}$. This allows the model to capture the full spatial context, which is crucial for preserving the topological characteristics of 3D tubular shapes during both training and inference. During training, the whole image region $\mathcal{X}$ and its corresponding ground-truth mask $\mathcal{Y}$ are provided as input, and the network parameters $\theta$ are optimized by minimizing a segmentation loss function $\mathcal{L}_{\text{seg}}$ (e.g., Dice loss):
\begin{equation}\label{eq::seg}
\theta^{*} = \arg\min_{\theta} \ \mathcal{L}_{\text{seg}}(\mathcal{F}(\mathcal{X}; \theta), \mathcal{Y}).
\end{equation}
When the training process finished, the model with the parameters of $\theta^{*}$ can directly infer the entire image region during testing, producing a segmentation mask $\hat{\mathcal{Y}} = \mathcal{F}(\mathcal{X}; \theta)$ without the need for patch-wise aggregation. 

\subsection{Topological Integrity Betti}
Topological Integrity Betti (TIB) combines Betti number guidance with topological integrity preservation to refine predictions during the test stage. 
As seen in Fig.\ref{fig:TopoSculpt_Pipeline}.b), 
once $\mathcal{F}$ is trained, it produces initial segmentations on unseen test images, which are then iteratively refined by TIB ($\mathcal{L}_{\text{TIB}}$). 
Generally speaking, the iterative refinement process is formulated as: $\theta^{i+1} := \theta^{i} - \eta \, \nabla_{\theta} \mathcal{L}_{\text{TIB}}\bigl(\theta^{(i)}\bigr), \; i \geq 0, \; \theta^{0} = \theta^{*}$.
The framework consists of two components, $\mathcal{L}_{\text{TIB}_{cor}}$ and $\mathcal{L}_{\text{TIB}_{com}}$, responsible for correcting topological errors and preserving overall integrity, respectively.
The correction term $\mathcal{L}_{\text{TIB}_{cor}}$ leverages known topological priors encoded by Betti numbers. Given the initial probability map $\hat{\mathcal{Y}}$ produced by the trained network $\mathcal{F}_{\theta^{*}}$, where $\hat{\mathcal{Y}} =   \mathcal{F}_{\theta^{*}}(\mathcal{X}) ,\hat{\mathcal{Y}} : \mathbb{R}^{3} \rightarrow [0, 1]$, persistent homology (PH) \cite{otter2017roadmap, zomorodian2005topology} is applied to characterize the evolving topology of $\hat{\mathcal{Y}}$ across all possible thresholds $p$. This process yields a dynamic description of the probabilistic segmentation topology and its critical events. The prior topology term ($Topo_{P}$) can then be incorporated without additional annotation effort; for instance, the pulmonary airway is known to satisfy $\beta_{0} = 1$, corresponding to a single connected component. 
Since the ground-truth topology $\mathcal{Y}$ is deterministic and its PH barcode remains persistent across all thresholds, we distinguish between faithful and superfluous features when analyzing the evolving topology of $\hat{\mathcal{Y}}$. The most persistent feature is defined as the faithful persistent feature ($Topo_{F}$), whereas all other features are regarded as superfluous persistent features ($Topo_{S}$).
Accordingly, the correction term $\mathcal{L}_{\text{TIB}_{cor}}$ is designed to encourage the persistence of faithful features while suppressing spurious ones. It is formulated as:
\begin{equation}
 \mathcal{L}_{\text{TIB}_{cor}} = (Topo_{P} - Topo_{F}^{i} + Topo_{S}^{i}), \; i \geq 0,
\end{equation}
where $Topo_{P}$ denotes the topology prior, $Topo_{F}^{i}$ represents the faithful feature, and $Topo_{S}^{i}$ corresponds to the superfluous feature at the $i$-th refinement step generated from $\hat{\mathcal{Y}^{i}} = \mathcal{F}(\mathcal{X}; \theta^{i})$. 
$Topo_{P}$ is a constant persistent feature derived from the known topology of the target structure, while $Topo_{F}^{i}$ and $Topo_{S}^{i}$ are dynamic persistent features that evolve with the segmentation during each refinement iteration.
As illustrated in Fig.\ref{fig:TopoSculpt_Framework}.a.2), enforcing only $\mathcal{L}_{\text{TIB}_{cor}}$ may induce shortcut learning, which leads to substantial structural distortions and undermines topological integrity. To mitigate this issue, we introduce the integrity preservation term $\mathcal{L}_{\text{TIB}_{com}}$, designed to penalize deviations from the original topology of the predicted segmentation. This term is defined from two complementary perspectives,voxel-wise constraint and structural-wise constraint, as follows:

\begin{align}
\mathcal{L}_{\text{TIB}_{com}} 
&= \alpha \times \frac{1}{N} \left|\mathcal{F}_{\theta^{i+1}}(\mathcal{X}) 
- \mathcal{F}_{\theta^{i}}(\mathcal{X})\right|^{2} \notag \\
&\quad - \beta \times 2 \times 
\frac{Struc_{prec} \times Struc_{sens}}
{ Struc_{prec} + Struc_{sens}},
\end{align}

where the first term serves as a voxel-wise constraint by minimizing the mean squared error between consecutive predictions, with $N$ denoting the total number of voxels in $\mathcal{X}$. The second term acts as a structural-wise constraint by maximizing structural similarity throughout refinement, thereby promoting the recovery of topological integrity rather than its degradation, as illustrated in Fig.\ref{fig:TopoSculpt_Framework}.a.3) and further detailed in Fig.~\ref{fig:TopoSculpt_Framework}.c). 
Inspired by \cite{shit2021cldice}, we quantify the proposed structural similarity using structural precision ($Struc_{prec}$) and structural sensitivity ($Struc_{sens}$), which are computed from the soft skeletons of the predicted segmentations weighted by the likelihood maps across two consecutive iterations.

\begin{align}
\mathrm{Struc_{prec}}(SP^{i+1}, VP^{i}) 
&= \frac{|SP^{i+1} \cap VP^{i}|}{|SP^{i+1}|}, \\[6pt]
\mathrm{Struc_{sens}}(SP^{i}, VP^{i+1}) 
&= \frac{|SP^{i} \cap VP^{i+1}|}{|SP^{i}|}.
\end{align}

Unlike \cite{shit2021cldice}, we do not have the access to ground-truth skeletons during test time refinement. Therefore, we compute two consecutive soft skeletons weighted by their respective likelihood maps to approximate the structural similarity, which are defined as:
\begin{equation}
\begin{aligned}
SP^{i+1} &= \operatorname{\mathrm{soft-skel}}\!\left( \mathcal{F}_{\theta^{i+1}}(\mathcal{X}) > 0.5 \right) \odot \mathcal{F}_{\theta^{i+1}}(\mathcal{X}), \\
SP^{i} &= \operatorname{\mathrm{soft-skel}}\!\left( \mathcal{F}_{\theta^{i}}(\mathcal{X}) > 0.5 \right) \odot \mathcal{F}_{\theta^{i}}(\mathcal{X}),\\
VP^{i+1} &= \mathcal{F}_{\theta^{i+1}}(\mathcal{X}), \, \, VP^{i} = \mathcal{F}_{\theta^{i}}(\mathcal{X}).
\end{aligned}
\end{equation}
In summary, the overall TIB loss ($\mathcal{L}_{\text{TIB}}$) is formulated as:
\begin{equation}
\mathcal{L}_{\text{TIB}} = \mathcal{L}_{\text{TIB}_{cor}} + \mathcal{L}_{\text{TIB}_{com}}.
\end{equation}  

\subsection{Curriculum Refinement}
It is well known that persistent homology (PH) calculation for whole 3D structures is computationally expensive \cite{otter2017roadmap, zomorodian2005topology, bauer2021ripser}. We observed that Betti numbers change rapidly in the early refinement stages but stabilize in later stages, as illustrated in Fig.~\ref{fig:TopoSculpt_Framework}.b). To mitigate the computational burden, we design a progressive refinement paradigm that gradually shifts the focus from large-scale to fine-scale topological errors. Accordingly, we initially compute PH features densely with fewer steps to correct large-scale errors, and then progressively increase the sampling interval to concentrate on finer-scale corrections. Specifically, the curriculum refinement is defined as:


\begin{equation}
\begin{aligned}
\mathcal{L}_{\text{TIB}} &= (Topo_{P} - Topo_{F}^{i} + Topo_{S}^{i}) 
+ \mathcal{L}_{\text{TIB}_{com}}, \\
&\quad 0 \leq i \leq t, \\
\mathcal{L}_{\text{TIB}} &= (Topo_{P} - Topo_{F}^{j(i)} + Topo_{S}^{j(i)}) 
+ \gamma \mathcal{L}_{\text{TIB}_{com}}, \\
&\quad t \leq i \leq T, \quad j(i) = \left\lfloor \tfrac{i}{k} \right\rfloor \cdot k.
\end{aligned}
\end{equation}

When the refinement enters the later stages ($i > t$), Betti numbers become more stable and large-scale corrections are less frequent. To reduce computational cost, we introduce a sampling mechanism through the function $j(i) = \lfloor i/k \rfloor \cdot k$, which selects representative iterations spaced by interval $k$. At these sampled steps, the topological prior $Topo_{P}$ is enforced, while the integrity term $\mathcal{L}_{\text{TIB}_{com}}$ is down-weighted by $\gamma$ to avoid over-regularization.This design enables TopoSculpt to first perform dense and aggressive topological correction, and then progressively shift towards sparse and fine-scale adjustments, thereby achieving both efficiency and accuracy in topological refinement.

\section{Experiments and Results}

\makeatletter
\def\hlinew#1{%
\noalign{\ifnum0=`}\fi\hrule \@height #1 \futurelet
\reserved@a\@xhline}
\makeatother
\begin{table*}[!t]
\renewcommand\arraystretch{1.7}
\caption{Quantitative results on binary modeling on ATM'22 pulmonary airway, TopCoW24 Circle of Willis (CoW), and computed tomography coronary angiography (CTCA). The 
centerlineDice (clDice, \%), Normalized Surface Dice (NSDice, \%), 
95\% Hausdorff distance (HD, $mm$), Branch detected rate (BD, \%), 
Tree length detected rate (TD, \%), and Betti-0 error ($\beta_{0}$) are reported. 
HVModel denotes the holistic-view model.
} 
\label{tab:singleclass_all}
\centering
\scalebox{1.0}{
\begin{tabular}{>{\centering\arraybackslash}p{1.5cm}|>{\centering\arraybackslash}p{3.4cm}|>{\centering\arraybackslash}p{1.6cm}>{\centering\arraybackslash}p{1.6cm}>{\centering\arraybackslash}p{1.6cm}>{\centering\arraybackslash}p{1.6cm}>{\centering\arraybackslash}p{1.6cm}>{\centering\arraybackslash}p{1.6cm}}
\hlinew{1pt}
\textbf{Dataset} & \textbf{Method}  & \textbf{clDice} $\uparrow$ & \textbf{NSDice} $\uparrow$ & \textbf{HD} $\downarrow$ & \textbf{BD} $\uparrow$ & \textbf{TD} $\uparrow$ & \textbf{$\beta_{0}$ error} $\downarrow$ \\ 
\hlinew{0.5pt}
\multirow{10}{*}{ATM'22}
& Dice+CE  &   86.27\scriptsize{$\pm$3.44}  &  95.20\scriptsize{$\pm$2.20}  & 1.56\scriptsize{$\pm$1.04}  & 73.25\scriptsize{$\pm$8.52}   & 74.65\scriptsize{$\pm$7.83}   & 96.40\scriptsize{$\pm$27.40}  \\
& clDice \cite{shit2021cldice}         &  88.81\scriptsize{$\pm$4.50}  &   91.69\scriptsize{$\pm$2.36}  &  4.92\scriptsize{$\pm$7.31}   &  85.55\scriptsize{$\pm$8.31}   &   85.63\scriptsize{$\pm$7.78}   &   102.10\scriptsize{$\pm$34.80}               \\
& CAL  \cite{zhang2023towards}          &  89.64\scriptsize{$\pm$3.19}  &   97.51\scriptsize{$\pm$3.19}  &  1.29\scriptsize{$\pm$1.16}   &  85.23\scriptsize{$\pm$7.69}   &   85.58\scriptsize{$\pm$7.41}   &   103.03\scriptsize{$\pm$32.07}               \\
& SkeletonRecall \cite{kirchhoff2024skeleton}   &  87.69\scriptsize{$\pm$3.46}  &   96.57\scriptsize{$\pm$1.78}  & 3.00\scriptsize{$\pm$5.77}   & 87.52\scriptsize{$\pm$5.26}    &  87.88\scriptsize{$\pm$4.95} &        104.33\scriptsize{$\pm$36.15}    \\
& cbDice \cite{shi2024centerline}         &  85.31\scriptsize{$\pm$6.09}  &  94.65\scriptsize{$\pm$3.18} & 2.65\scriptsize{$\pm$4.33}     &  80.63\scriptsize{$\pm$11.46}  &  81.81\scriptsize{$\pm$11.01}                  &    64.27\scriptsize{$\pm$23.62}               \\
& BoundaryLoss \cite{kervadec2019boundary}  &  88.23\scriptsize{$\pm$4.67}     &  97.40\scriptsize{$\pm$1.90}   &  1.58\scriptsize{$\pm$2.88}  &  80.08\scriptsize{$\pm$10.18}   &   80.23\scriptsize{$\pm$9.29}  & 102.87\scriptsize{$\pm$34.24}    \\
& TopoLoss \cite{hu2019topology}  &  75.14\scriptsize{$\pm$30.87}    &   86.99\scriptsize{$\pm$25.52}  &  40.56\scriptsize{$\pm$47.99}    &   69.17\scriptsize{$\pm$31.28}      &   69.58\scriptsize{$\pm$30.54}  &   114.10\scriptsize{$\pm$49.67}    \\
& DMT \cite{hu2021topology}  &   75.54\scriptsize{$\pm$32.91}  &  83.01\scriptsize{$\pm$28.97}  & 37.59\scriptsize{$\pm$46.76}  &  69.50\scriptsize{$\pm$32.42}  & 69.69\scriptsize{$\pm$31.92}  & 111.70\scriptsize{$\pm$53.02}       \\
\cline{2-8}
& HVModel  &   90.48\scriptsize{$\pm$2.92}  &  96.96\scriptsize{$\pm$1.50}  & 1.07\scriptsize{$\pm$0.56}  & 82.06\scriptsize{$\pm$7.44}   & 82.23\scriptsize{$\pm$6.37}   & 69.00\scriptsize{$\pm$19.01}  \\
& TopoSculpt & \textbf{92.27\scriptsize{$\pm$1.91}}  & \textbf{98.11\scriptsize{$\pm$0.82}}  & \textbf{1.02\scriptsize{$\pm$0.29}} & \textbf{91.33\scriptsize{$\pm$4.84}}  &  \textbf{91.56\scriptsize{$\pm$4.09}} &  \textbf{3.40\scriptsize{$\pm$2.44}} \\ 
\hlinew{0.5pt}
\multirow{10}{*}{TopCoW24}
& Dice+CE  &   90.39\scriptsize{$\pm$4.68}  &  94.87\scriptsize{$\pm$4.16}  & 3.34\scriptsize{$\pm$3.73}  & 68.06\scriptsize{$\pm$17.14}   & 64.47\scriptsize{$\pm$17.50}   & 5.55\scriptsize{$\pm$3.90}  \\
& clDice \cite{shit2021cldice}  &  93.78\scriptsize{$\pm$3.24}  &  96.17\scriptsize{$\pm$3.59}  & 1.89\scriptsize{$\pm$1.72} &  79.07\scriptsize{$\pm$16.15}  & 74.08\scriptsize{$\pm$17.29} & 1.80\scriptsize{$\pm$1.57} \\
& CAL  \cite{zhang2023towards}  &  93.69\scriptsize{$\pm$3.21}  &  96.06\scriptsize{$\pm$3.62}  & 1.91\scriptsize{$\pm$1.68} &  80.48\scriptsize{$\pm$15.98}  & 78.21\scriptsize{$\pm$16.82} & 2.25\scriptsize{$\pm$1.51} \\
& SkeletonRecall \cite{kirchhoff2024skeleton} & 92.80\scriptsize{$\pm$3.87}  &  96.09\scriptsize{$\pm$3.71}  & 1.75\scriptsize{$\pm$1.64} &  79.12\scriptsize{$\pm$18.22}  & 76.42\scriptsize{$\pm$19.40} & 2.80\scriptsize{$\pm$1.81} \\
& cbDice \cite{shi2024centerline} & 91.87\scriptsize{$\pm$4.03}  &  95.90\scriptsize{$\pm$3.47}  & 2.03\scriptsize{$\pm$1.70} &  63.66\scriptsize{$\pm$13.65}  & 59.01\scriptsize{$\pm$12.83} & 2.25\scriptsize{$\pm$1.37} \\
& BoundaryLoss \cite{kervadec2019boundary}  &  92.96\scriptsize{$\pm$3.08}  &  95.91\scriptsize{$\pm$3.27}  & 1.97\scriptsize{$\pm$1.67} &  73.33\scriptsize{$\pm$17.28} & 70.72\scriptsize{$\pm$16.92}  &  2.80\scriptsize{$\pm$1.83} \\
& TopoLoss \cite{hu2019topology}  & 91.19\scriptsize{$\pm$4.31}  &  95.90\scriptsize{$\pm$3.47}  & 2.03\scriptsize{$\pm$1.70} &  63.66\scriptsize{$\pm$13.65}  & 59.01\scriptsize{$\pm$12.83} & 2.25\scriptsize{$\pm$1.37} \\
& DMT \cite{hu2021topology}  & 90.94\scriptsize{$\pm$4.03} & 95.27\scriptsize{$\pm$3.44} & 2.51\scriptsize{$\pm$1.92}  & 71.67\scriptsize{$\pm$15.97} &  67.85\scriptsize{$\pm$17.32} &  5.00\scriptsize{$\pm$2.79} \\
\cline{2-8}
& HVModel  & 96.67\scriptsize{$\pm$1.98}  &  99.10\scriptsize{$\pm$1.67}  & 0.73\scriptsize{$\pm$0.69}   & 84.26\scriptsize{$\pm$16.67}   & 82.84\scriptsize{$\pm$17.21}  & 1.65\scriptsize{$\pm$1.19}   \\
& TopoSculpt &  \textbf{97.98\scriptsize{$\pm$1.00}}   &  \textbf{99.26\scriptsize{$\pm$1.22}} & \textbf{0.62\scriptsize{$\pm$0.39}}   & \textbf{96.72\scriptsize{$\pm$8.62}}   & \textbf{96.73\scriptsize{$\pm$8.62}} & \textbf{0.30\scriptsize{$\pm$0.56}}  \\ 
\hlinew{0.5pt}
\multirow{10}{*}{CTCA}
& Dice+CE  &  84.09\scriptsize{$\pm$4.54}  &  93.18\scriptsize{$\pm$4.12}  & 6.35\scriptsize{$\pm$5.67}  & 71.76\scriptsize{$\pm$12.86}   & 70.49\scriptsize{$\pm$14.35}   & 33.50\scriptsize{$\pm$10.34}  \\
& clDice \cite{shit2021cldice}  &  85.82\scriptsize{$\pm$3.54}  &  95.01\scriptsize{$\pm$2.66}  & 3.59\scriptsize{$\pm$2.71} &  68.52\scriptsize{$\pm$13.60}  & 70.74\scriptsize{$\pm$10.58} & 14.75\scriptsize{$\pm$4.35} \\
& CAL  \cite{zhang2023towards}  &  85.48\scriptsize{$\pm$2.35}  &  94.56\scriptsize{$\pm$2.55}  & 4.52\scriptsize{$\pm$3.44} &  67.39\scriptsize{$\pm$13.21}  & 71.38\scriptsize{$\pm$13.59} & 18.12\scriptsize{$\pm$5.80} \\
& SkeletonRecall \cite{kirchhoff2024skeleton} & 85.08\scriptsize{$\pm$3.07}  &  94.20\scriptsize{$\pm$2.69}  & 4.72\scriptsize{$\pm$3.67} & 77.86\scriptsize{$\pm$13.53}  & 80.71\scriptsize{$\pm$12.22} & 16.50\scriptsize{$\pm$6.96} \\
& cbDice \cite{shi2024centerline} & 80.07\scriptsize{$\pm$5.35}  &  91.94\scriptsize{$\pm$4.10}  & 7.16\scriptsize{$\pm$4.81} &  56.96\scriptsize{$\pm$12.29}  & 59.34\scriptsize{$\pm$15.26} & 11.00\scriptsize{$\pm$3.43} \\
& BoundaryLoss \cite{kervadec2019boundary}  &82.68\scriptsize{$\pm$6.10}  &  94.63\scriptsize{$\pm$2.29}  & 4.45\scriptsize{$\pm$2.70} &  54.65\scriptsize{$\pm$11.18}  & 63.29\scriptsize{$\pm$15.49} & 13.88\scriptsize{$\pm$5.73} \\
& TopoLoss \cite{hu2019topology}  & 84.26\scriptsize{$\pm$5.27}  &  94.18\scriptsize{$\pm$2.91}  & 4.15\scriptsize{$\pm$2.48} &  63.93\scriptsize{$\pm$13.56}  & 64.11\scriptsize{$\pm$15.18} & 19.62\scriptsize{$\pm$7.87} \\
& DMT \cite{hu2021topology}  &85.16\scriptsize{$\pm$4.68}  &  94.30\scriptsize{$\pm$2.95}  & 4.40\scriptsize{$\pm$2.99} &  70.12\scriptsize{$\pm$12.81}  & 72.82\scriptsize{$\pm$14.80} & 24.25\scriptsize{$\pm$10.30} \\
\cline{2-8}
& HVModel  & 87.27\scriptsize{$\pm$6.89}  &  95.84\scriptsize{$\pm$5.93}  & 0.78\scriptsize{$\pm$0.58}   & 75.17\scriptsize{$\pm$7.65}   & 78.98\scriptsize{$\pm$10.37}  & 8.75\scriptsize{$\pm$6.16}   \\
& TopoSculpt &  \textbf{90.01\scriptsize{$\pm$3.53}}   &  \textbf{97.68\scriptsize{$\pm$2.66}} & \textbf{0.58\scriptsize{$\pm$0.26}}   & \textbf{91.33\scriptsize{$\pm$5.56}}   & \textbf{91.68\scriptsize{$\pm$5.91}} & \textbf{0.88\scriptsize{$\pm$1.36}}  \\ 
\hlinew{1pt}
\end{tabular}}
\end{table*}

\begin{figure*}[!t]
\centering
\includegraphics[width=1.00\linewidth]{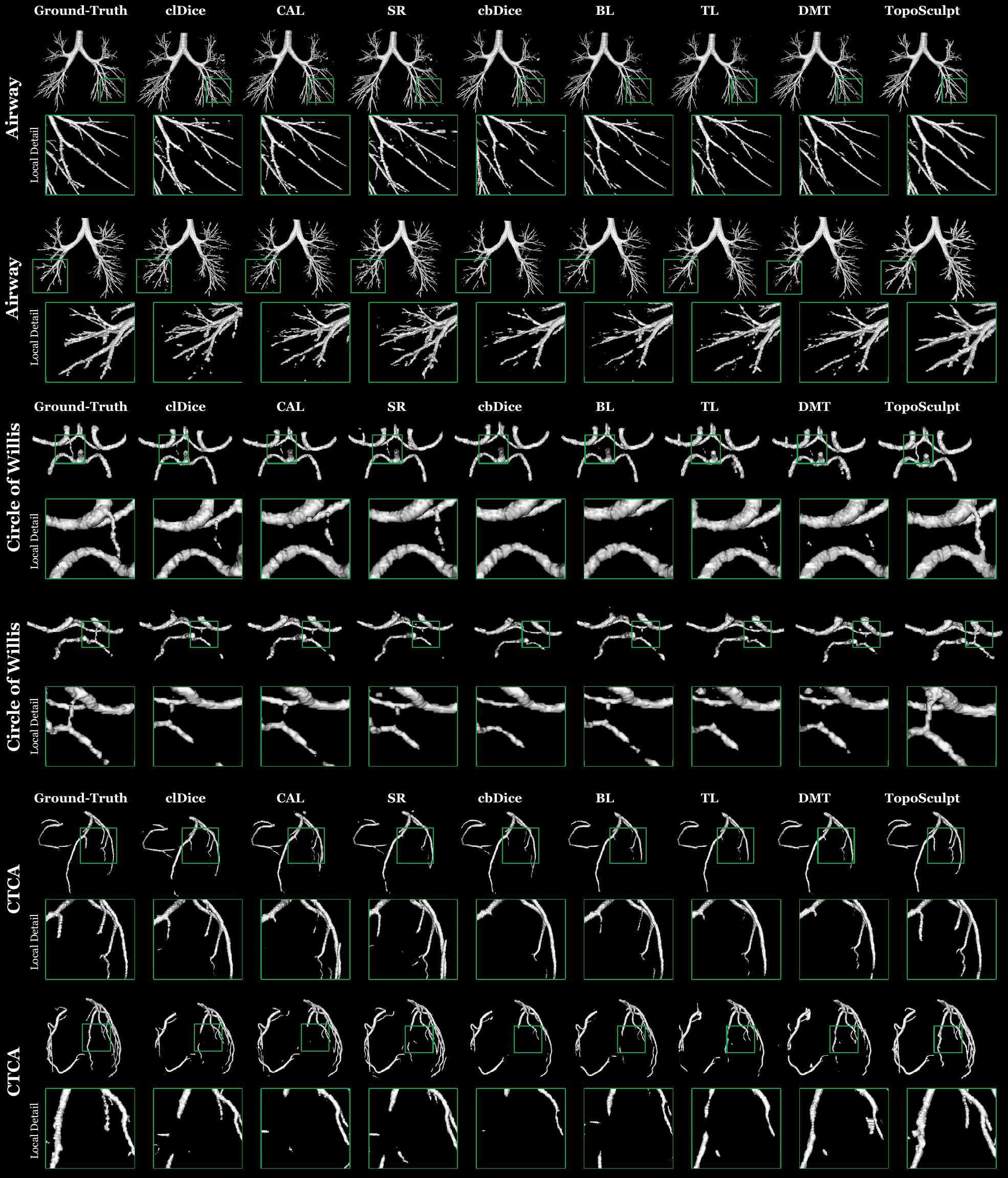}
\caption{Qualitative comparison of TopoSculpt with competing methods, including clDice, CAL, SkeletonRecall (SR), cbDice, BoundaryLoss (BL), TopoLoss (TL) and DMT, on three representative fine-grained tubular datasets. The green boxes highlight local regions where TopoSculpt better preserves fine structural details and achieves higher topological fidelity.}
\label{fig:fig_6}
\end{figure*}

\begin{figure*}[!t]
\centering
\includegraphics[width=1.0\linewidth]{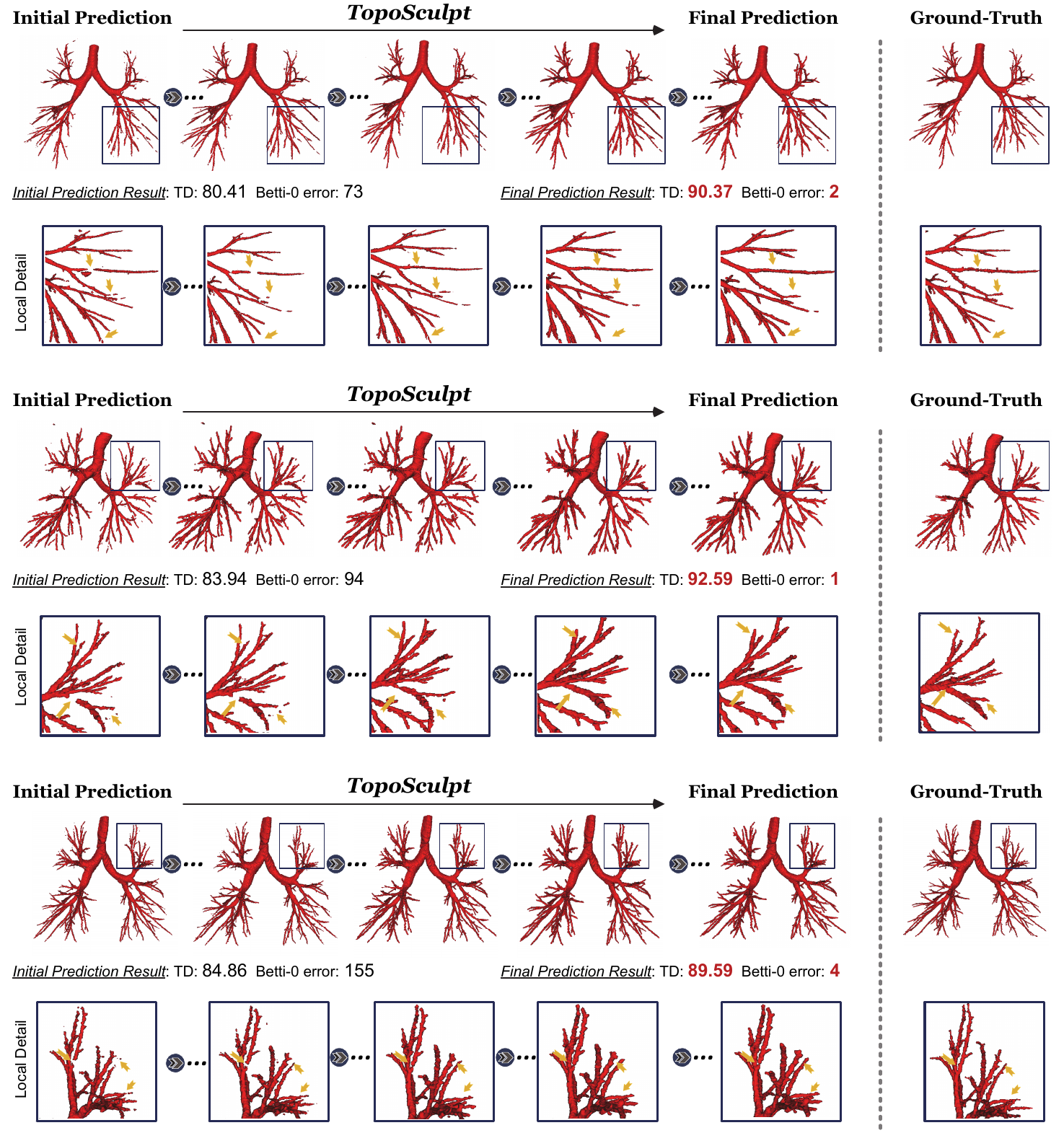}
\caption{The visualization of TopoSculpt illustrates the refinement process from the initial airway structure to the final result, exhibiting substantial enhancement in topological fidelity.
In each case, the first row depicts the holistic view of the airway structures, demonstrating progressive improvements in overall topology. The second row offers a zoomed-in perspective, emphasizing localized refinements and enhanced topological consistency in the detailed airway branches.}
\label{fig:Visual_Result_AirwaySingleCls}
\end{figure*}

\begin{figure*}[!t]
\centering
\includegraphics[width=0.9\linewidth]{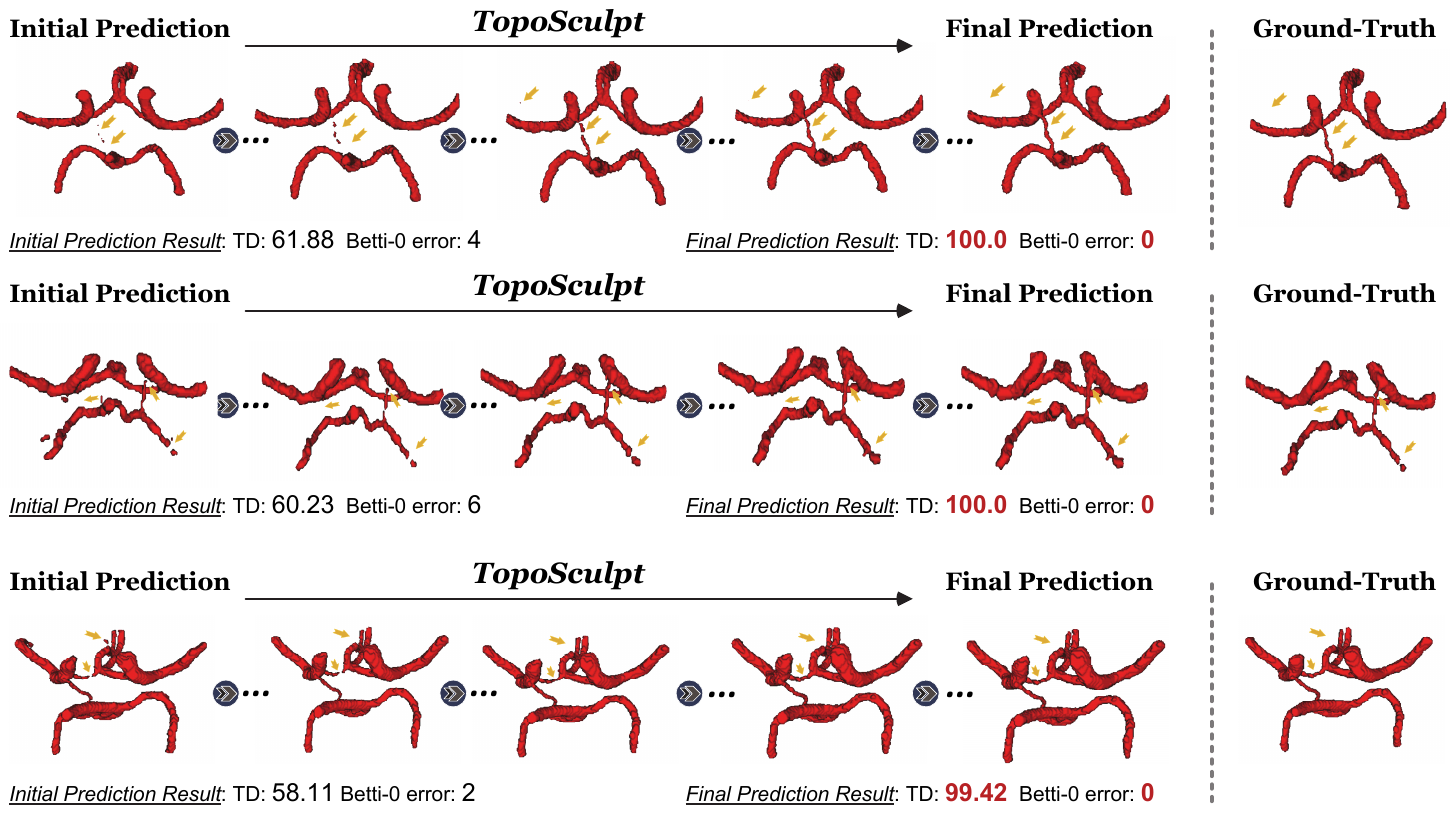}
\caption{The visualization of TopoSculpt illustrates the refinement process from the initial Circle of Willis (CoW) structure to the final result, exhibiting substantial enhancement in topological fidelity. The arrows indicate the locations of topological errors in the initial segmentation, which are effectively corrected by TopoSculpt.}
\label{fig:Visual_Result_TopCoWSingleCls}
\end{figure*}

\begin{figure*}[!t]
\centering
\includegraphics[width=0.9\linewidth]{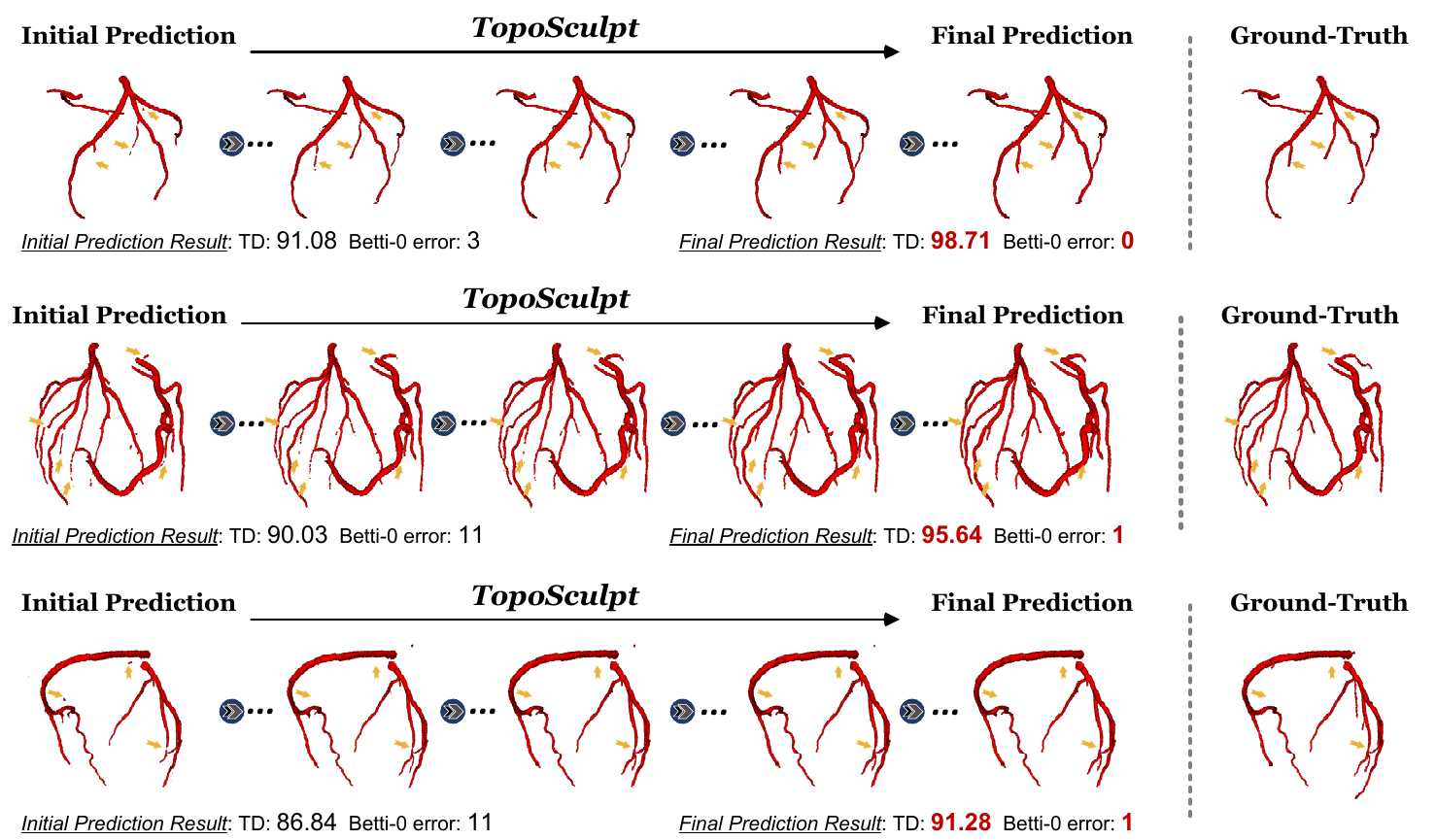}
\caption{The visualization of TopoSculpt illustrates the refinement process from the initial coronary  structure to the final result, exhibiting substantial enhancement in topological fidelity. The arrows indicate the locations of topological errors in the initial segmentation, which are effectively corrected by TopoSculpt.}
\label{fig:Visual_Result_ASOCASingleCls}
\end{figure*}
\begin{figure*}[!t]
\centering
\includegraphics[width=1.0\linewidth]{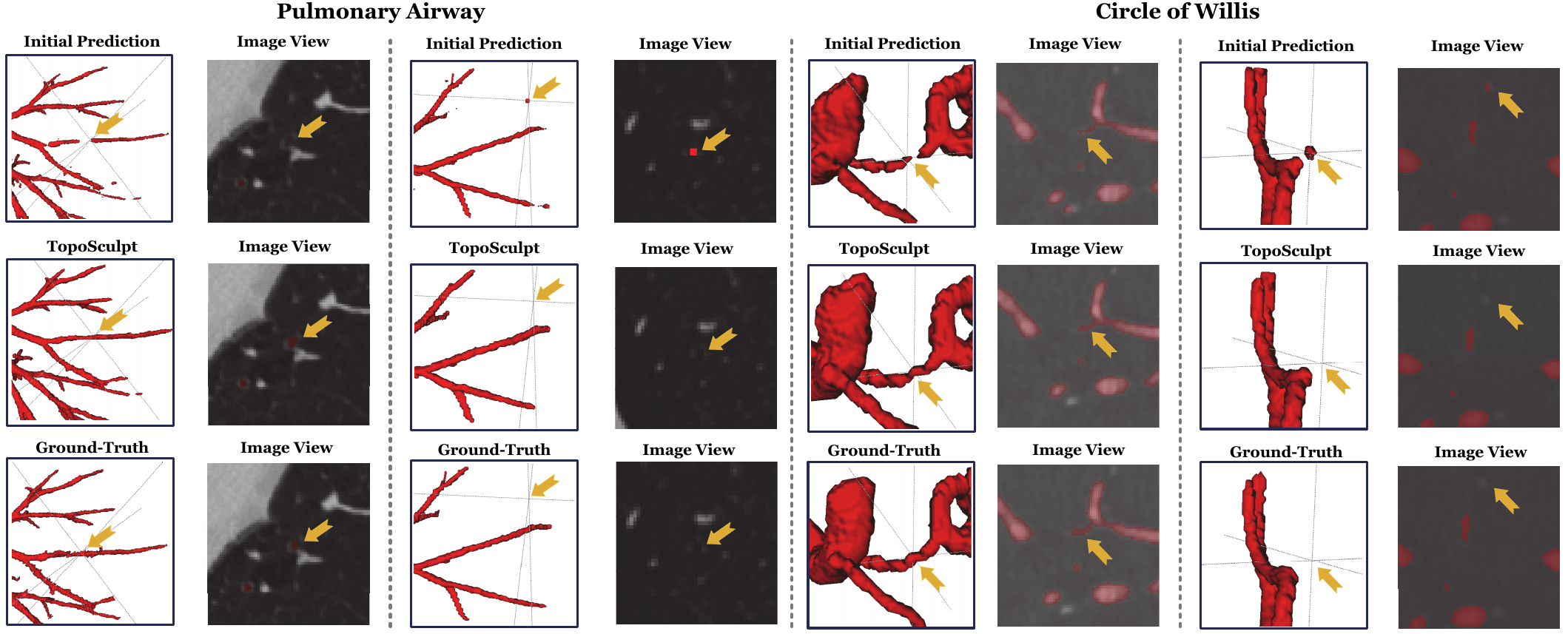}
\caption{Qualitative comparisons of topological refinement results on the pulmonary airway (left) and Circle of Willis (right) datasets. Each block shows the 3D structure (left) and the corresponding image view (right). The initial predictions (top row) exhibit typical topological errors, such as branch discontinuities, spurious components, or breakages (highlighted by arrows). After refinement with TopoSculpt (middle row), these errors are effectively suppressed, leading to restored connectivity and improved structural integrity that more closely matches the ground truth (bottom row). The results confirm that TopoSculpt not only eliminates erroneous Betti-0 components but also preserves the fine-scale geometry of complex tubular structures across different anatomical regions.}
\label{fig:Visual_Result_ImagePatch}
\end{figure*}

\begin{figure*}[!t]
\centering
\includegraphics[width=1.0\linewidth]{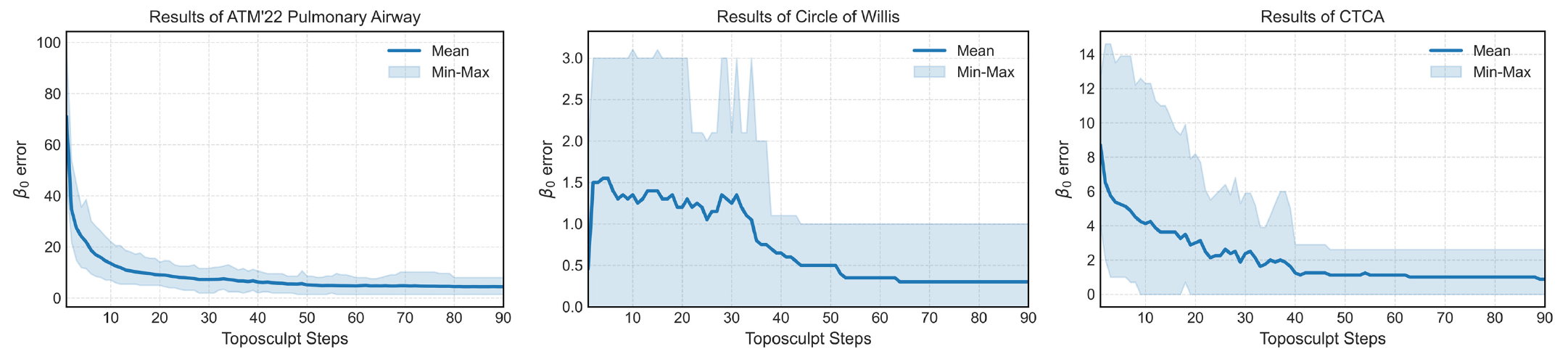}
\caption{Evolution of $\beta_{0}$ error during the TopoSculpt refinement process across three representative datasets: ATM'22 Pulmonary Airway (left), Circle of Willis (middle), and CTCA (right). The solid line denotes the mean $\beta_{0}$ error, while the shaded region indicates the min-max range across all test cases. We observed that Betti numbers change rapidly in the early refinement stages, reflecting the correction of large-scale connectivity errors, but gradually stabilize in later stages as the structures converge toward topologically consistent configurations. This trend confirms the effectiveness of the progressive refinement strategy in efficiently restoring and preserving global topological integrity.}
\label{fig:CurriculumProcess}
\end{figure*}

\begin{figure*}[!t]
\centering
\includegraphics[width=1.0\linewidth]{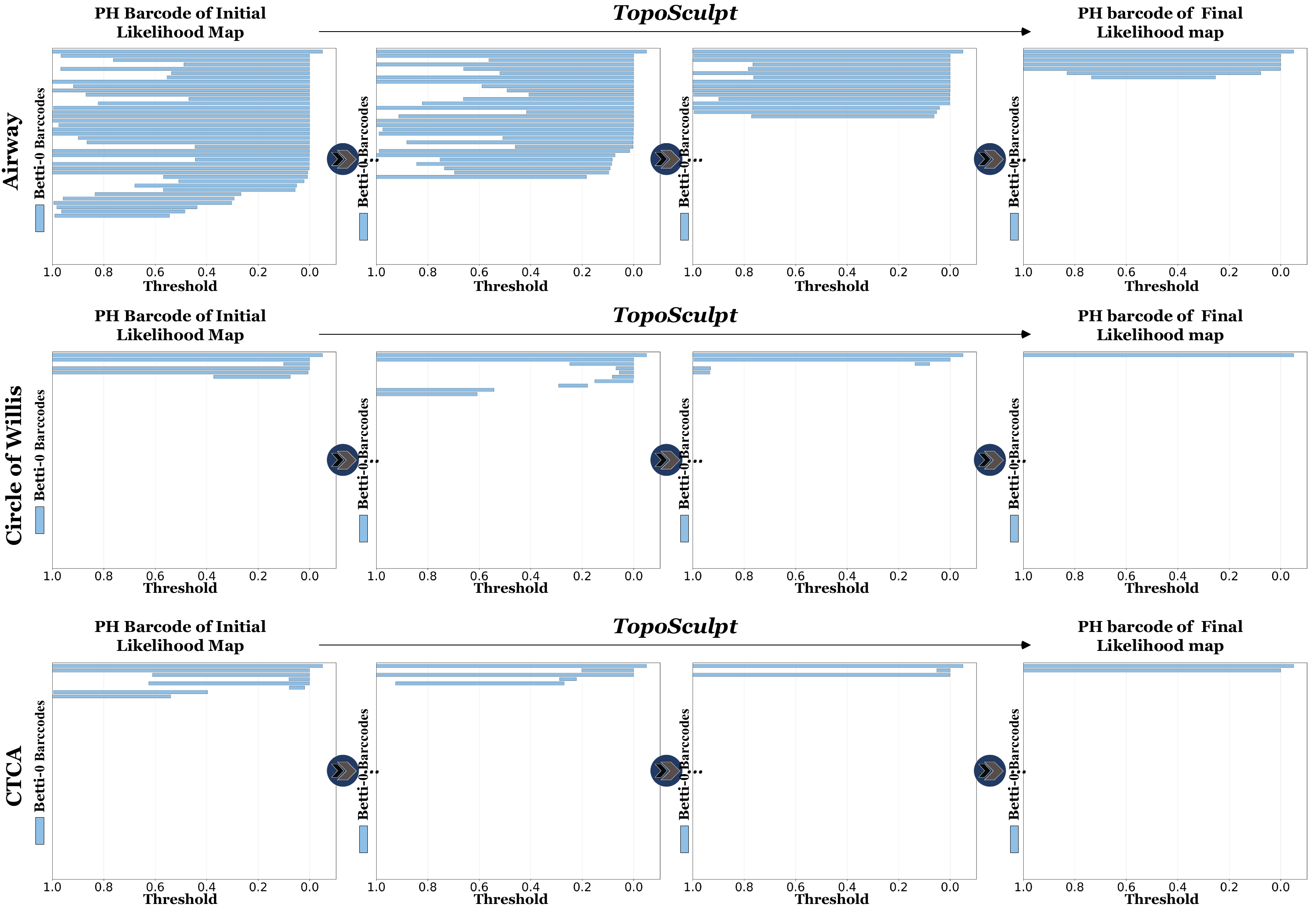}
\caption{Evolution of PH barcodes of the likelihood map during the TopoSculpt refinement process across three representative datasets: ATM'22 Pulmonary Airway (top), Circle of Willis (middle), and CTCA (bottom). Each horizontal line in the barcode represents a topological component feature and its persistence over threshold variations.}
\label{fig:PH_Barcode_Evolution_of_Likelihoodmap}
\end{figure*}

\begin{figure}[!t]
\centering
\includegraphics[width=1.0\linewidth]{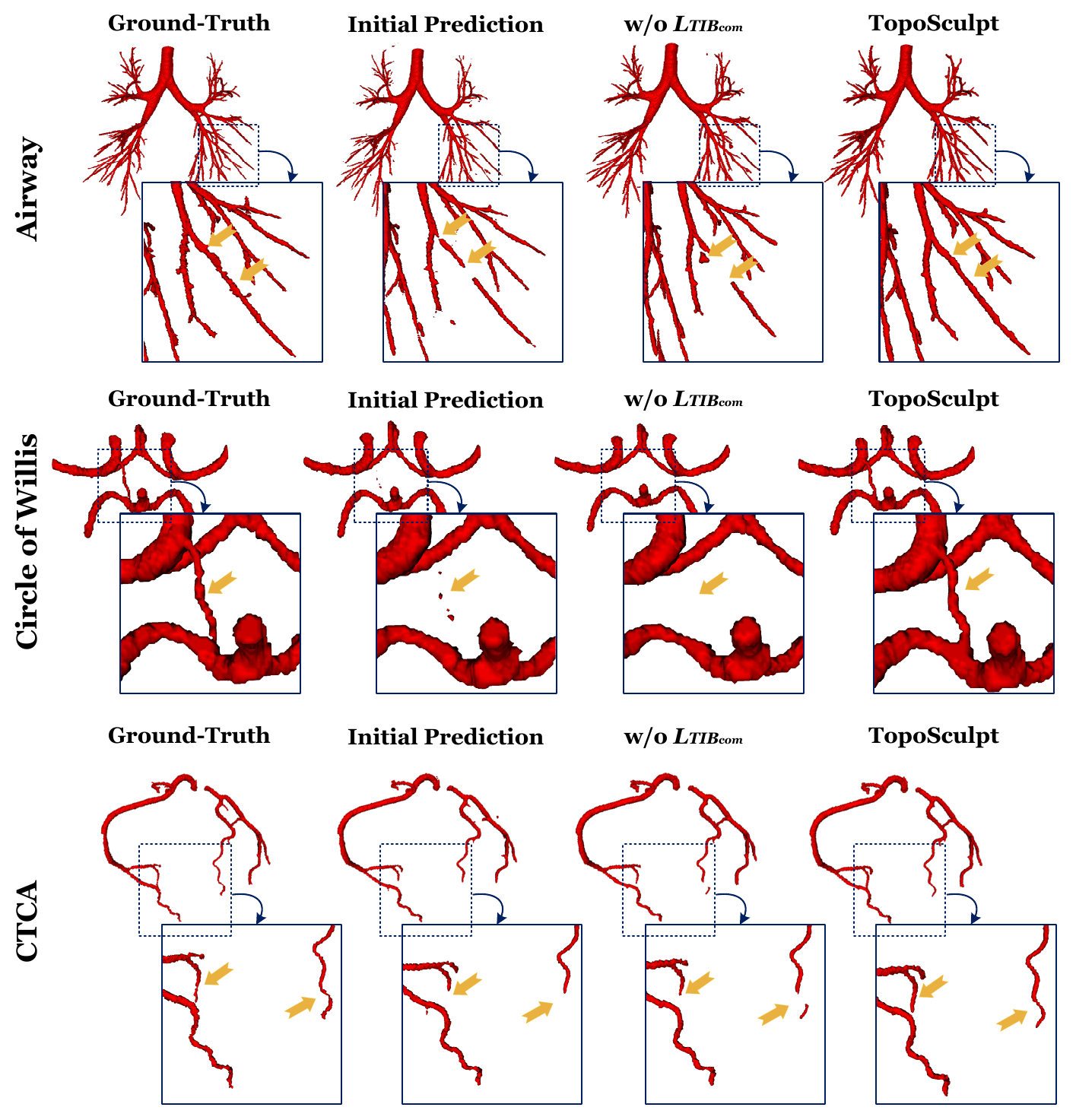}
\caption{Ablation study on the $\text{TIB}_{com}$ component of the proposed TopoSculpt framework across three representative datasets, ATM’22 Pulmonary Airway (top), Circle of Willis (middle), and CTCA (bottom). 
$\text{TIB}_{com}$ demonstrates its contribution to improving topological integrity and structural consistency.}
\label{fig:Ablation_TIB}
\end{figure}

\subsection{Dataset Information}
We evaluate TopoSculpt on three challenging benchmarks of three-dimensional tubular anatomy: \\
\textbf{ATM'22 Pulmonary Airway Dataset}\cite{zhang2023multi,zhang2024airmorph}. This dataset comprises multi-site, multi-domain chest CT scans with detailed airway tree annotations. The airway labels capture complex branching structures, frequent bifurcations, and substantial variability in branch radii. For our experiments, we selected a subset of 150 CT scans, allocating 120 scans for training and validation and 30 scans for testing. Each CT scan is first clamped to the range of [-1000, 400] HU, followed by normalization to [0, 1].\\
\textbf{TopCoW'24 Circle of Willis (CoW) Dataset} \cite{yang2024benchmarking}. The Circle of Willis is a critical vascular network that provides collateral blood supply between the anterior and posterior circulations of the brain. In this study, we focused on structures satisfying the topological prior $\beta_{0}=1$. Accordingly, we selected 71 CTA cases from the TopCoW24 dataset, with 51 cases used for training and validation and 20 cases reserved for testing. Each CT scan is first clamped to the range of [-1000, 1800] HU, followed by normalization to [0, 1].\\
\textbf{ASOCA'23 Computed Tomography Coronary Angiography (CTCA) Dataset} \cite{gharleghi2023annotated}. CTCA contains 40 cases of computed tomography coronary angiography images with annotations of the coronary artery tree. The coronary artery tree is a complex vascular structure with numerous fine branches and high anatomical variability. We randomly selected 32 cases for training and validation and 8 cases for testing. Each CT scan is first clamped to the range of [-100, 500] HU, followed by normalization to [0, 1].

\subsection{Implementation Details}
For implementation, the holistic-view model (HVModel) in Eq.\eqref{eq::seg} is built upon a standard 3D U-Net architecture \cite{cciccek20163d} with an initial channel width of 16, operating directly on the full image region. For comparison, we additionally implement a patch-wise variant of the same 3D U-Net where a patch size of $128 \times 128 \times 128$ is used for all experiments. The patch-wise network is trained with multiple losses designed for tubular structure segmentation. Recent topology-aware losses, including clDice \cite{shit2021cldice}, CAL \cite{zhang2023towards}, SkeletonRecall \cite{kirchhoff2024skeleton}, cbDice \cite{shi2024centerline}, BoundaryLoss \cite{kervadec2019boundary}, TopoLoss \cite{hu2019topology}, and DMT \cite{hu2021topology}, were included for comparison.

During the refinement stage of TopoSculpt, we set $\alpha = 1e4$ and $\beta = 1e3$ in the $\mathcal{L}_{\text{TIB}_{com}}$, and for the curriculum refinement strategy, we set $t=30$, $T=90$, $k=3$, and $\gamma=0.1$. AdamW \cite{loshchilov2017decoupled} optimizer is used with an learning rate of $1e-3$ for pulmonary airway, $1e-4$ for Circle of Willis and $1e-4$ for CTCA. All experiments are conducted on two NVIDIA A800 GPUs each with 80GB memory.

\subsection{Quantitative Results Analysis}\label{sec::quantitative_results}

The quantitative performance of TopoSculpt and competing methods was evaluated on three challenging 3D tubular structure datasets, with comprehensive results presented in TALBE \ref{tab:singleclass_all}. Several observations are listed as follows:

\textbf{Holistic vs Patch-wise}: A consistent finding across all datasets is the superior performance of our holistic-view (HVModel) baseline compared to the standard patch-wise approach trained with the combination of Dice Loss and CrossEntropy Loss (denoted by \textbf{Dice+CE} in TALBE \ref{tab:singleclass_all}), particularly in topological metrics. For instance, the $\beta_{0}$ error is reduced from $96.40$ to $69.00$ on the airway dataset, from $5.55$ to $1.65$ on the CoW dataset, and from $33.50$ to $8.75$ on the CTCA dataset, underscoring the advantage of global context modeling. Further, with the proposed TopoSculpt refinement based on HVModel, we observe substantial improvements across all datasets.

\textbf{ATM22 Dataset}:  As shown in TABLE \ref{tab:singleclass_all}, the most challenging ATM'22 pulmonary airway dataset clearly demonstrates the superiority of TopoSculpt. While most state-of-the-art methods struggled severely with maintaining topological correctness, often yielding $\beta_{0}$ errors exceeding 100 (e.g., CAL \cite{zhang2023towards} and SkeletonRecall \cite{kirchhoff2024skeleton}), TopoSculpt drastically reduced this error from the HVModel baseline of 69.00 to only 3.40. This fundamental correction in topological integrity directly translated into state-of-the-art structural completeness, with BD improving to 91.33\% and TD to 91.56\%. Such performance markedly surpasses strong competitors, including CAL (85.23\% BD) and SkeletonRecall (87.52\% BD). Moreover, PH-based approaches such as TopoLoss \cite{hu2019topology} and DMT \cite{hu2021topology} failed to perform well on this dataset, exhibiting clDice scores around 75\% alongside high HD/Betti errors. In contrast, TopoSculpt achieved the highest geometric fidelity, reaching 92.27\% in clDice, while simultaneously ensuring topological completeness.

\textbf{TopCow24 Dataset}: Consistent improvement was observed on the TopCoW'24 Circle of Willis dataset, as illustrated in TABLE \ref{tab:singleclass_all}. TopoSculpt achieved high topological completeness with a BD of 96.72\% and TD of 96.73\%, alongside a clDice of 97.98\%. This was a remarkable improvement from the HVModel's BD of 84.26\% and TD of 82.84\%. 
TopoSculpt effectively eliminated Betti errors by reducing the $\beta_{0}$ error from 1.65 to 0.30. 
In contrast, all other SOTA methods, including clDice \cite{shit2021cldice} and CAL \cite{zhang2023towards}, failed to exceed 81\% in BD or TD, highlighting the superiority of TopoSculpt in correcting critical vascular disconnection.

\textbf{CTCA Dataset}: On the CTCA coronary angiography dataset, as seen in TABLE \ref{tab:singleclass_all}, TopoSculpt significantly outperformed all other methods. It demonstrated superior topological correctness, reducin the $\beta_{0}$ error to 0.88 from the baseline's 8.75, while many other methods struggled to reduce this error below 10. This improvement in topological integrity was accompanied by notable gains in completeness metrics, with TD increasing from 78.98\% to 91.68\% (+12.7\%) and BD from 75.17\% to 91.33\% (+16.2\%). It can be observed that TopoSculpt surpassed all other SOTA methods by a large margin. For instance, the next best competitor, SkeletonRecall, only achieved a TD of 80.71\%. As for overlap-based geometric accuracy, TopoSculpt achieved a clDice of 90.01\%, 
NSDice of 97.68\%, and BD of 91.33\%. This represents a substantial improvement over the HVModel baseline, which scored 87.27\% clDice and only 75.17\% BD.

In summary, the quantitative results across all three datasets confirm that TopoSculpt not only improves upon its holistic-view baseline but also decisively outperforms all competing SOTA methods. It uniquely demonstrates the ability to correct severe topological errors (lowering $\beta_{0}$) and, crucially, translates these corrections into comprehensive and significant improvements in both structural completeness (TD, BD) and overlap-wise   (clDice, NSDice).

\subsection{Qualitative Results Analysis}
Fig.\ref{fig:fig_6} presents a direct visual comparison between TopoSculpt and several competing methods across all three datasets. The results clearly show that TopoSculpt consistently achieves superior structural integrity and greater consistency to the ground truth. The local regions highlighted by green boxes further emphasize TopoSculpt's ability to preserve fine structural details and maintain topological fidelity, whereas other methods frequently exhibit breakages, discontinuities, or spurious noise in these areas.

Fig.\ref{fig:Visual_Result_AirwaySingleCls}, Fig.\ref{fig:Visual_Result_TopCoWSingleCls}, and Fig.\ref{fig:Visual_Result_ASOCASingleCls} 
visualize the iterative refinement process of TopoSculpt on unseen test cases. Unlike conventional methods where model parameters are fixed post-training, leaving them unable to correct topological inconsistencies on a case-by-case basis during inference, TopoSculpt performs a test-time refinement supervised by a topological integrity prior. These figures visually confirm that each refinement step progressively improves the reconstruction, steering the initial prediction towards a topologically correct and structurally complete result. 
Concretely, for the pulmonary airway (Fig.\ref{fig:Visual_Result_AirwaySingleCls}), the initial prediction often suffer from severe fragmentation, 
resulting in high $\beta_0$ errors (e.g., 73, 94, 155) and low TD rates. 
The figure illustrates TopoSculpt's progressive refinement, showing how it systematically reconnects fragmented subtrees and resolves broken links. The final predictions exhibit dramatically reduced $\beta_0$ errors (e.g., 2, 1, 4) and correspondingly high TD scores. The local detail views confirm the restoration of continuity in peripheral branches. For the CoW dataset, initial predictions exhibit critical disconnections at key arterial junctions, reflected in low 
topological integrity and non-zero $\beta_0$ errors. TopoSculpt is shown to effectively correct these errors , leading to anatomically faithful vascular networks with higher TD scores satisfactory and $\beta_0$ numbers in the final predictions. Similarly, on the CTCA dataset, the initial predictions contain topological errors. The refinement process is shown to effectively resolve these, reducing the betti errors and significantly boosting TD scores. In Fig.~\ref{fig:Visual_Result_ImagePatch}, we provide qualitative results that illustrate how TopoSculpt corrects critical topological inconsistencies, shown in both 3D reconstructions and the corresponding image patch views. For the fine-grained pulmonary airway and CoW structures, the initial predictions often exhibit breakages or spurious isolated components, which directly correspond to erroneous $\beta_{0}$ features, as seen in the first row of Fig.~\ref{fig:Visual_Result_ImagePatch}, pointed out by arrows. Such defects compromise structural connectivity and consequently lower branch and tree detection rates. By contrast, TopoSculpt effectively suppresses these Betti errors, restoring continuous branches while eliminating false positives. Importantly, the refinement is achieved without inducing geometric distortions, thereby preserving local morphology and lumen boundaries, as illustrated in the second row of Fig.~\ref{fig:Visual_Result_ImagePatch}. This demonstrates the complementary strength of TopoSculpt: it explicitly enhances topological integrity while simultaneously maintaining geometric fidelity, yielding reconstructions that are both anatomically consistent and visually coherent with the ground truth.
Additional supplementary results could be found here: \url{https://github.com/Puzzled-Hui/TopoSculpt}.

\makeatletter
\def\hlinew#1{%
\noalign{\ifnum0=`}\fi\hrule \@height #1 \futurelet
\reserved@a\@xhline}
\makeatother
\begin{table*}[!t]
\renewcommand\arraystretch{1.7}
\caption{Ablation of the effectiveness of topological integrity $\mathcal{L}_{\text{TIB}_{com}}$ in TopoSculpt. 
clDice (\%), NSDice (\%), HD (mm), Branch length detected rate (BD, \%), 
Tree length detected rate (TD, \%), and Betti-0 error ($\beta_{0}$) are reported. 
}
\label{tab:ablationtopointegrity}
\centering
\scalebox{1.0}{
\begin{tabular}{>{\centering\arraybackslash}p{3.8cm}>{\centering\arraybackslash}p{1.6cm}>{\centering\arraybackslash}p{1.6cm}>{\centering\arraybackslash}p{1.6cm}>{\centering\arraybackslash}p{1.6cm}>{\centering\arraybackslash}p{1.6cm}>{\centering\arraybackslash}p{1.6cm}}
\hlinew{1pt}
\multicolumn{7}{c}{\textbf{ATM'22 Pulmonary Airway}}\\\hlinew{0.25pt}
\textbf{Method}  & \textbf{clDice} $\uparrow$ & \textbf{NSDice} $\uparrow$ & \textbf{HD} $\downarrow$ & \textbf{BD} $\uparrow$ & \textbf{TD} $\uparrow$ & \textbf{$\beta_{0}$ error} $\downarrow$ \\ \hlinew{0.5pt}
Toposculpt w/o $\mathcal{L}_{\text{TIB}_{com}}$     & 89.74\scriptsize{$\pm$2.55}  & 96.88\scriptsize{$\pm$1.07} &  1.41\scriptsize{$\pm$0.75}  & 83.92\scriptsize{$\pm$5.63}   &  84.48\scriptsize{$\pm$5.29}   & 12.36\scriptsize{$\pm$6.33}    \\
Toposculpt & \textbf{92.27\scriptsize{$\pm$1.91}}  & \textbf{98.11\scriptsize{$\pm$0.82}}  & \textbf{1.02\scriptsize{$\pm$0.29}} & \textbf{91.33\scriptsize{$\pm$4.84}}  &  \textbf{91.56\scriptsize{$\pm$4.09}} &  \textbf{3.40\scriptsize{$\pm$2.44}} \\ \hlinew{0.5pt}
\multicolumn{7}{c}{\textbf{Circle of Willis (CoW)}}\\\hlinew{0.25pt}
\textbf{Method}  & \textbf{clDice} $\uparrow$ & \textbf{NSDice} $\uparrow$ & \textbf{HD} $\downarrow$ & \textbf{BD} $\uparrow$ & \textbf{TD} $\uparrow$ & \textbf{$\beta_{0}$ error} $\downarrow$ \\ \hlinew{0.5pt}
Toposculpt w/o $\mathcal{L}_{\text{TIB}_{com}}$     &  96.92\scriptsize{$\pm$1.56} & 98.32\scriptsize{$\pm$6.10}  & 0.71\scriptsize{$\pm$0.27} & 91.58\scriptsize{$\pm$14.20} & 90.79\scriptsize{$\pm$14.86}  & 0.45\scriptsize{$\pm$0.67}      \\
TopoSculpt &  \textbf{97.98\scriptsize{$\pm$1.00}}   &  \textbf{99.26\scriptsize{$\pm$1.22}} & \textbf{0.62\scriptsize{$\pm$0.39}}   & \textbf{96.72\scriptsize{$\pm$8.62}}   & \textbf{96.73\scriptsize{$\pm$8.62}} & \textbf{0.30\scriptsize{$\pm$0.56}} \\ \hlinew{0.5pt}
\multicolumn{7}{c}{\textbf{Computed Tomography Coronary Angiography (CTCA)}}\\\hlinew{0.25pt}
\textbf{Method}  & \textbf{clDice} $\uparrow$ & \textbf{NSDice} $\uparrow$ & \textbf{HD} $\downarrow$ & \textbf{BD} $\uparrow$ & \textbf{TD} $\uparrow$ & \textbf{$\beta_{0}$ error} $\downarrow$ \\ \hlinew{0.5pt}
TopoSculpt w/o $\mathcal{L}_{\text{TIB}_{com}}$     & 89.06\scriptsize{$\pm$2.55} & 96.83\scriptsize{$\pm$2.59}  & 0.69\scriptsize{$\pm$0.27}   &   88.17\scriptsize{$\pm$5.51}     &   90.43\scriptsize{$\pm$4.61} & \textbf{0.75\scriptsize{$\pm$0.83}}   \\
TopoSculpt &  \textbf{90.01\scriptsize{$\pm$3.53}}   &  \textbf{97.68\scriptsize{$\pm$2.66}} & \textbf{0.58\scriptsize{$\pm$0.26}}   & \textbf{91.33\scriptsize{$\pm$5.56}}   & \textbf{91.68\scriptsize{$\pm$5.91}} & 0.88\scriptsize{$\pm$1.36}  \\ 
\hlinew{1pt}
\end{tabular}}
\end{table*}

\makeatletter
\def\hlinew#1{%
\noalign{\ifnum0=`}\fi\hrule \@height #1 \futurelet
\reserved@a\@xhline}
\makeatother
\begin{table*}[!t]
\renewcommand\arraystretch{1.7}
\caption{Comparison of different configurations on ATM'22 pulmonary airway. 
clDice (\%), NSDice (\%), HD95 (mm), Branch length detected rate (BD, \%), 
Tree length detected rate (TD, \%), and Betti-0 error ($\beta_{0}$) are reported.}
\label{tab:config_comparison}
\centering
\scalebox{0.85}{
\begin{tabular}{>{\centering\arraybackslash}p{3.4cm}
                >{\centering\arraybackslash}p{2.5cm}
                >{\centering\arraybackslash}p{1.5cm}
                >{\centering\arraybackslash}p{1.7cm}
                >{\centering\arraybackslash}p{1.7cm}
                >{\centering\arraybackslash}p{1.7cm}
                >{\centering\arraybackslash}p{1.7cm}
                >{\centering\arraybackslash}p{1.7cm}
                >{\centering\arraybackslash}p{1.7cm}}
\hlinew{1pt}
\textbf{Method} & \textbf{Patchsize} & \textbf{Params (M)} &
\textbf{clDice} $\uparrow$ & \textbf{NSDice} $\uparrow$ & \textbf{Hd95 (mm)} $\downarrow$ &
\textbf{BD} $\uparrow$ & \textbf{TD} $\uparrow$ & \textbf{$\beta_{0}$ error}  $\downarrow$ \\ 
\hlinew{0.5pt}
3D U-Net   & $128\times128\times128$ & 4.083   &
86.27\scriptsize{$\pm$3.44} & 95.20\scriptsize{$\pm$2.20} & 1.56\scriptsize{$\pm$1.04} &
73.25\scriptsize{$\pm$8.52} & 74.65\scriptsize{$\pm$7.83} & 96.40\scriptsize{$\pm$27.40} \\
3D U-Net   & Holistic View           & 4.083   &
90.48\scriptsize{$\pm$2.92} & 96.96\scriptsize{$\pm$1.50} & \underline{1.07\scriptsize{$\pm$0.56}} &
\underline{82.06\scriptsize{$\pm$7.44}} & 82.23\scriptsize{$\pm$6.37} & 69.00\scriptsize{$\pm$19.01} \\
nnUNet    & $160\times128\times112$  & 30.786 &
\underline{90.50\scriptsize{$\pm$3.77}} & \underline{97.52\scriptsize{$\pm$1.60}} & \underline{1.14\scriptsize{$\pm$1.21}} &
\underline{83.88\scriptsize{$\pm$8.69}} & \underline{84.26\scriptsize{$\pm$7.86}} & \underline{43.77\scriptsize{$\pm$13.04}} \\
nnUNet    & $64\times64\times64$    & 30.786 &
88.59\scriptsize{$\pm$3.51} & 96.55\scriptsize{$\pm$2.30} & 4.00\scriptsize{$\pm$9.73} &
81.22\scriptsize{$\pm$8.50} & 82.10\scriptsize{$\pm$8.021} & 79.10\scriptsize{$\pm$20.96} \\
nnUNet    & $128\times128\times128$ & 30.786 &
90.39\scriptsize{$\pm$3.82} & 97.17\scriptsize{$\pm$3.45} & 1.28\scriptsize{$\pm$1.86} &
83.47\scriptsize{$\pm$9.98} & 83.78\scriptsize{$\pm$9.54} & 49.70\scriptsize{$\pm$14.50} \\
3D U-Net w/ TopoSculpt& Holistic View           & 4.083   &
\textbf{92.27\scriptsize{$\pm$1.91}} & \textbf{98.11\scriptsize{$\pm$0.82}} & \textbf{1.02\scriptsize{$\pm$0.29}} &
\textbf{91.33\scriptsize{$\pm$4.84}} & \textbf{91.56\scriptsize{$\pm$4.09}} & \textbf{3.40\scriptsize{$\pm$2.44}} \\
\hlinew{1pt}
\end{tabular}}
\end{table*}

\subsection{Ablation Study}
\noindent\textbf{Visualization of PH barcode variation.} 
Fig.\ref{fig:PH_Barcode_Evolution_of_Likelihoodmap} visualizes the evolution of persistent homology (PH) barcodes for the likelihood maps during the refinement process of TopoSculpt across all three datasets. Each horizontal line in the barcode represents a topological component feature and its persistence over threshold variations. These visualizations clearly demonstrate that TopoSculpt not only reduces the overall $\beta_{0}$ error but also promotes topological stability in the PH space. In the initial predictions, a large number of short-lived PH bars are observed, indicating the presence of numerous spurious components and local discontinuities. As the refinement proceeds, TopoSculpt progressively suppresses these unstable Betti features while preserving only the dominant, persistent structures corresponding to the true anatomical topology. This trend is particularly evident in the airway dataset, where the dense clusters of short Betti-0 bars gradually diminish, converging to several long-lived components that reflects the enhancement of connectivity. Similar stabilization can be observed in the CoW and CTCA datasets, where the number of spurious bars rapidly decreases and the remaining features exhibit longer persistence and higher structural consistency. 

\noindent\textbf{Curriculum learning process.} 
Fig.\ref{fig:CurriculumProcess} illustrates the evolution of $\beta_{0}$ error throughout the TopoSculpt refinement process across three datasets. The solid line represents the mean $\beta_{0}$ error, while the shaded area denotes the min-max range across all test cases. 
A clear curriculum-like learning behavior can be observed. At the early refinement stages, the $\beta_{0}$ error decreases sharply, indicating that TopoSculpt efficiently eliminates large-scale connectivity errors through dense persistent homology (PH) sampling. As refinement continues, the rate of reduction gradually slows, and the curves stabilize, reflecting the transition from coarse-level correction to fine-grained structural refinement.
This trend confirms the effectiveness of the proposed progressive refinement scheme in improving both convergence efficiency and topological stability. By focusing on large-scale errors first and progressively shifting toward finer-scale corrections, TopoSculpt achieves efficient topological recovery while maintaining global structural integrity across diverse anatomical domains.

\noindent \textbf{Effectiveness of topological completeness constraint}.
By jointly optimizing $\mathcal{L}_{\text{TIB}_{cor}}$ and $\mathcal{L}_{\text{TIB}_{com}}$, TopoSculpt achieves a balanced refinement process that corrects topological errors while maintaining structural coherence. The qualitative results in Fig.\ref{fig:Ablation_TIB} clearly demonstrate this improvement. Without $\mathcal{L}_{\text{TIB}_{com}}$, the refined predictions exhibit residual disconnections in airway and coronary artery structures or missing vessel branches in CoW structures, as highlighted by the arrows. In contrast, the complete TopoSculpt framework produces continuous and topologically consistent reconstructions that are highly consistent with the ground truth. 
Quantitative results in TABLE \ref{tab:ablationtopointegrity} further validate the importance of $\mathcal{L}_{\text{TIB}_{com}}$. Across all three datasets, incorporating this term significantly improves both topological and geometric metrics. On the ATM'22 pulmonary airway dataset, $\mathcal{L}_{\text{TIB}_{com}}$ reduces the $\beta_{0}$ error from 12.36 to 3.40 and increases BD and TD from 83.92\% and 84.48\% to 91.33\% and 91.56\%, respectively. Similar improvements are observed for the CoW and CTCA datasets, where $\beta_{0}$ errors are further reduced and clDice and NSDice consistently increase. These results confirm that $\mathcal{L}_{\text{TIB}_{com}}$ is essential for preventing shortcut learning, preserving meaningful anatomical structures, and achieving globally consistent topological refinement across diverse three-dimensional tubular anatomies.

\noindent \textbf{Essential Role of TopoSculpt}.
TABLE~\ref{tab:config_comparison} presents a comprehensive comparison among TopoSculpt, its holistic-view baseline, and several nnUNet variants with substantially larger model capacities. Although adopting a holistic-view input yields improvements over patch-wise training, the resulting model still exhibits a high $\beta_{0}$ error of 69.00, indicating that global spatial context alone is insufficient to ensure reliable topological consistency. Despite achieving satisfactory clDice scores, all three nnUNet configurations demonstrate weak performance on BD, TD, and $\beta_{0}$ error, revealing that larger backbones provide limited capability in correcting connectivity defects in fine-grained tubular anatomy. 
In contrast, TopoSculpt achieves a substantial improvement while using the same lightweight backbone as the holistic-view baseline. It reduces the $\beta_{0}$ error from 69.00 to 3.40 and enhances both BD and TD by approximately 10\%, outperforming all larger nnUNet configurations by a considerable margin. These results indicate that TopoSculpt plays a critical role in restoring the topological fidelity of tubular structures, which cannot be attained through model scaling or holistic input alone.
The scaling behavior of large models on full-resolution volumetric data remains underexplored in this work. Nonetheless, a promising direction is to combine higher-capacity backbones with holistic full-volume processing and TopoSculpt-based refinement, particularly as high-memory GPUs and advanced computational resources become increasingly available.

\section{Discussion}
In this section, we discuss the clinical applications, limitations and potential future directions. 

The proposed TopoSculpt can be regarded as a scalable method or an extension to existing interactive foundation segmentation approaches. Beyond the conventional prompts used in current interactive methods\cite{ma2024segment,isensee2025nninteractive}, such as points, boxes, or scribbles, TopoSculpt could support the alternative geometric prior as the prompt to refine the initial segmentation results. 
A promising avenue for future work is the generation of geometric priors as language-formatted prompts and their integration, alongside TopoSculpt, into multimodal large language models. 

Although TopoSculpt performs zero-shot test-time refinement, it faces the same stopping-criteria challenge as other iterative methods. A meaningful future direction would be to design case-specific refinement strategies 
by defining the topological distance between the likelihood map and the expected geometric prior. Such a distance measure could serve as a quantitative indicator of how much refinement is required for each case, thereby providing an adaptive stopping criterion. This would allow the refinement process to terminate once the topological discrepancy has been minimized, ensuring both computational efficiency and reliable topological fidelity. 

TopoSculpt could also help alleviate the burden on clinicians when annotating fine tubular structures. As noted in airway datasets such as ATM22 and BAS, the provided labels are often incomplete. In such cases, the neural network may generate additional branches beyond a single connected component. Many of these branches, however, are clinically meaningful. By presenting them as candidate components, clinicians can rapidly review the results and filter out false positives. Then our proposed TopoSculpt can focus on refining the remaining structures. This process reduces manual labeling effort while ensuring both topological fidelity and geometrically accurate modeling.

\section{Conclusion}
In this work, we presented TopoSculpt, a holistic framework for refining three-dimensional tubular shape modeling by jointly leveraging Betti-guided constraints and topological integrity preservation. Unlike prior patch-wise or training-only approaches, TopoSculpt introduces whole-region modeling to ensure global topology consistency, while the Topological Integrity Betti (TIB) constraint balances correctness and structural fidelity. A curriculum refinement strategy further accelerates computation by progressively addressing errors across scales. Extensive experiments on pulmonary airways, the Circle of Willis, and coronary artery confirm that TopoSculpt achieves significant improvements in $\beta_{0}$ error reduction, tree and branch detection, as well as geometric overlap metrics. 


\bibliography{paper.bib}

\begin{IEEEbiography}[{\includegraphics[width=1in,height=1.25in,clip,keepaspectratio]{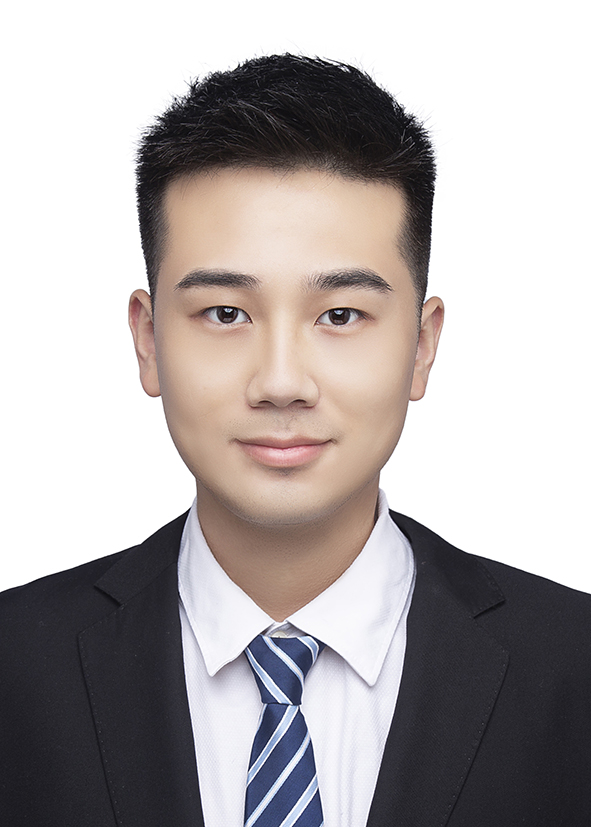}}]{Minghui Zhang}
received the B.S. degree in Department of Automation from Southeast University, Nanjing, China, in 2020. 
Currently, he is a Ph.D. student at the Department of Automation, Shanghai Jiao Tong University. 
He has authored several research papers in MICCAI/MedIA/IEEE JBHI. His major research is now focused on medical image analysis and deep learning.
\end{IEEEbiography}

\begin{IEEEbiography}[{\includegraphics[width=1in,height=1.25in,clip,keepaspectratio]{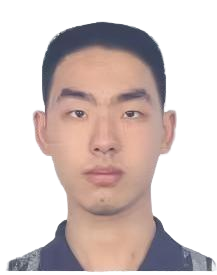}}]{Yaoyu Liu}
received the BS degree in electronic information and electrical engineering from Shanghai Jiao Tong University, Shanghai, China. 
He is currently working toward the PhD degree with School of Automation and Intelligent Sensing, Shanghai Jiao Tong University. 
His current research interests include medical image analysis, computer vision and artificial intelligence.
\end{IEEEbiography}

\begin{IEEEbiography}[{\includegraphics[width=1in,height=1.25in,clip,keepaspectratio]{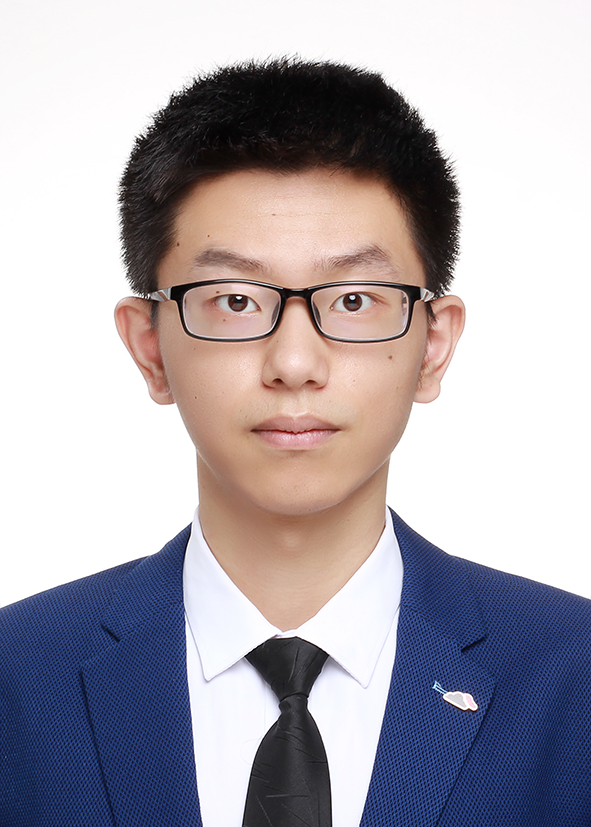}}]{Junyang Wu}
received the B.S. degree in biomedical engineering from Shanghai Jiao Tong University, Shanghai, China, in 2023. 
He is currently pursuing the Ph.D. degree with the department of biomedical engineering, Shanghai Jiao Tong University, Shanghai, China.
His current research interests include surgical robotics, medical image analysis, and deep learning.
\end{IEEEbiography}

\begin{IEEEbiography}[{\includegraphics[width=1in,height=1.25in,clip,keepaspectratio]{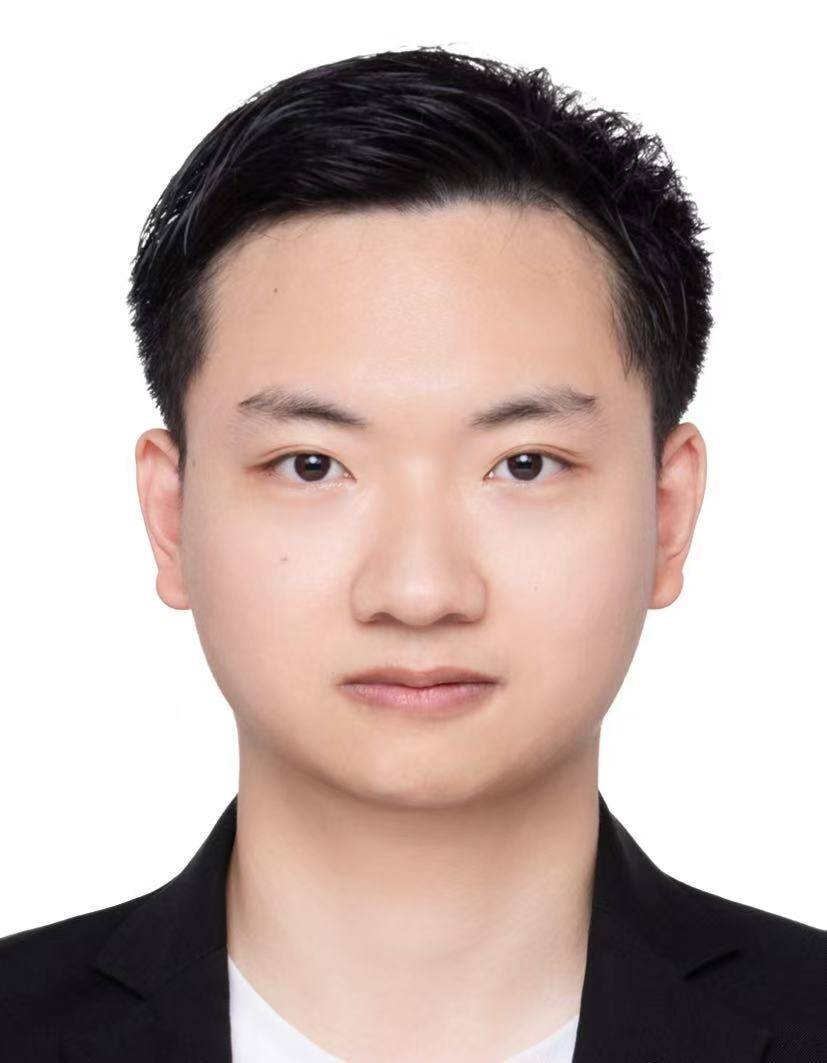}}]{Xin You}
received the B.S. degree in Department of Automation from Harbin Institute of Technology, Harbin, China, in 2020. 
He is currently working towards the Ph.D. degree majoring at the Institute of Image Processing and Pattern Recognition, Department of Automation, Shanghai Jiao Tong University, supervised by 
Prof. Yun Gu. His research interests include unified medical image segmentation, 
temporal motion modeling, medical image synthesis. He has published academic papers on IEEE TMI/JBHI/ICCV/MICCAI.
\end{IEEEbiography}

\begin{IEEEbiography}[{\includegraphics[width=1in,height=1.25in,clip,keepaspectratio]{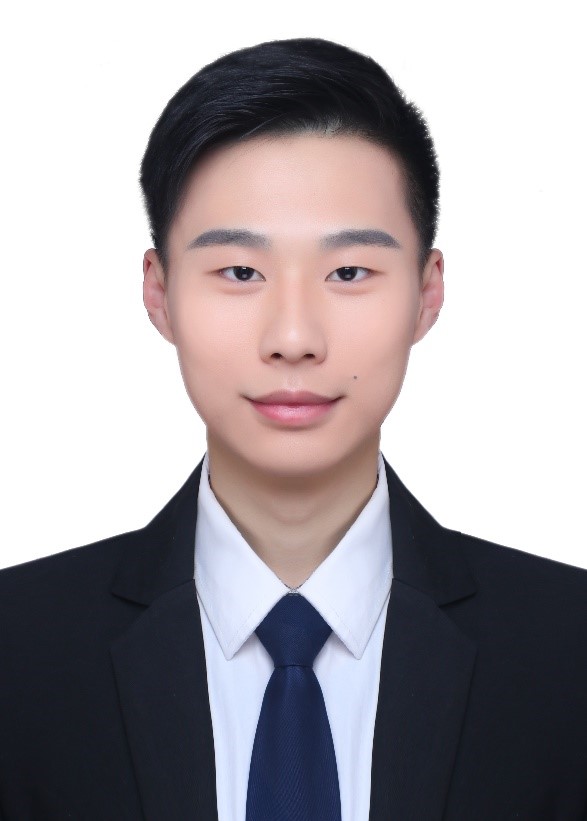}}]{Hanxiao Zhang}
received his master's degree from Imperial College London, UK. 
He obtained the Ph.D. in biomedical engineering under the supervision of Prof. Guang-Zhong Yang at Shanghai Jiao Tong University, China, 
and is currently a Postdoctoral Research Fellow at Shanghai Jiao Tong University. He is also affiliated with the Shanghai Key Laboratory of Flexible Medical Robotics. 
His research interests include medical image analysis, medical robotics, and computer-assisted intervention.
\end{IEEEbiography}

\begin{IEEEbiography}[{\includegraphics[width=1in,height=1.25in,clip,keepaspectratio]{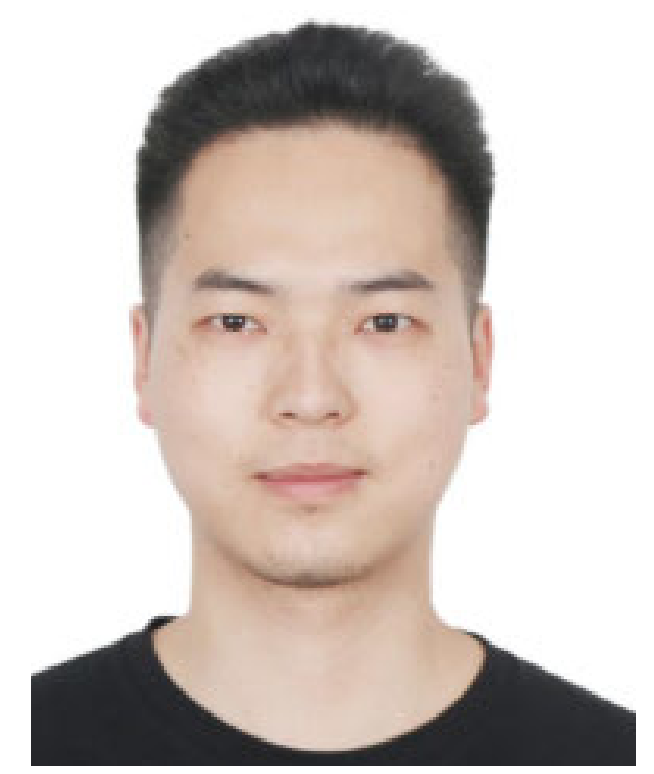}}]{Junjun He} is currently a researcher at Shanghai AI Laboratory, focusing on foundation models and general medical AI. He obtained the Ph.D. degree from Shanghai Jiao Tong University, China. He has published over 10 papers in international journals and conferences, including T-PAMI, 
TMI, CVPR, ICCV, ECCV, MICCAI, ISBI, etc. He was one of the main contributors of MMSegmentation. He won several championships in medical image 
computing competitions such as ODIR19, AutoPET22, and FLARE22.
\end{IEEEbiography}

\begin{IEEEbiography}[{\includegraphics[width=1in,height=1.25in,clip,keepaspectratio]{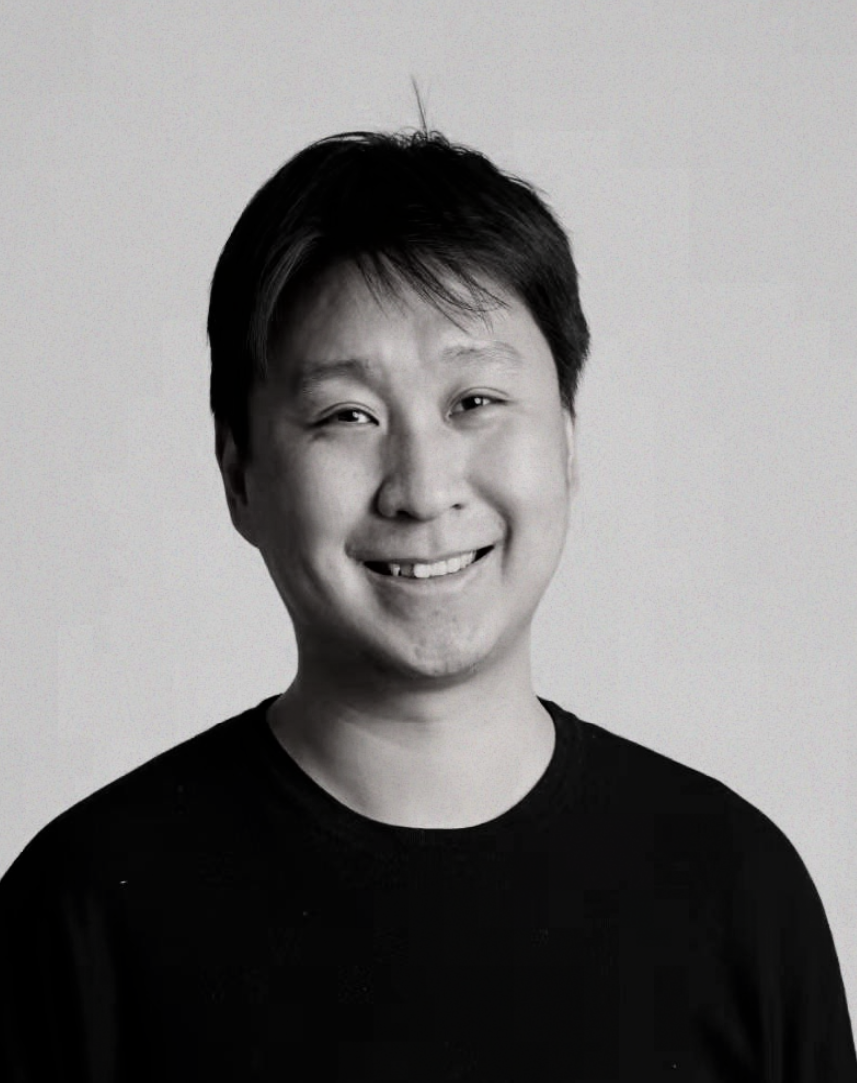}}]{Yun Gu} received the Ph.D. degree from the Department of Biomedical Engineering, Shanghai Jiao Tong University. 
He is currently an Assistant Professor with the Institute of Medical Robotics and the Institute of Image Processing and Pattern Recognition, Shanghai Jiao Tong University, China. 
He has published in major conferences and journals including MedIA/IEEE TMI/TIP/TNNLS/TBME/JBHI and MICCAI/ICRA/IROS. His major research interests are computer-assisted intervention 
and medical image analysis.
\end{IEEEbiography}

\begin{IEEEbiography}[{\includegraphics[width=1in,height=1.25in,clip,keepaspectratio]{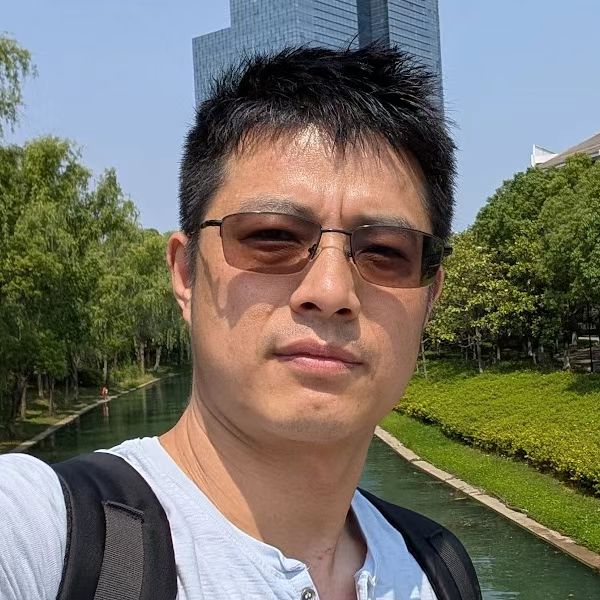}}]{Xinglin Zhang} is an AI researcher and practitioner specializing in medical imaging AI, especially large-scale foundation models, and multimodal modeling. He currently serves as the Chief Technology Officer (CTO) of Medical Image Insights and leads the development of the MIIA Medical Imaging Foundation Model, which is one of the largest and most extensively trained medical imaging foundation models to date, achieving state-of-the-art (SOTA) performance across multiple downstream imaging tasks and enabling practical, clinician-driven AI development (AI-DIY) in real clinical environment.
\end{IEEEbiography}

\end{document}